**PAPER**

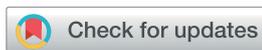 Check for updates

Cite this: DOI: 10.1039/d4dd00088a

# Uncertainty quantification for molecular property predictions with graph neural architecture search†

Shengli Jiang, [ID] *[a] Shiyi Qin,[a] Reid C. Van Lehn,[a] Prasanna Balaprakash[b] and Victor M. Zavala [ID] [ac]

Graph Neural Networks (GNNs) have emerged as a prominent class of data-driven methods for molecular property prediction. However, a key limitation of typical GNN models is their inability to quantify uncertainties in the predictions. This capability is crucial for ensuring the trustworthy use and deployment of models in downstream tasks. To that end, we introduce AutoGNNUQ, an automated uncertainty quantification (UQ) approach for molecular property prediction. AutoGNNUQ leverages architecture search to generate an ensemble of high-performing GNNs, enabling the estimation of predictive uncertainties. Our approach employs variance decomposition to separate data (aleatoric) and model (epistemic) uncertainties, providing valuable insights for reducing them. In our computational experiments, we demonstrate that AutoGNNUQ outperforms existing UQ methods in terms of both prediction accuracy and UQ performance on multiple benchmark datasets, and generalizes well to out-of-distribution datasets. Additionally, we utilize t-SNE visualization to explore correlations between molecular features and uncertainty, offering insight for dataset improvement. AutoGNNUQ has broad applicability in domains such as drug discovery and materials science, where accurate uncertainty quantification is crucial for decision-making.

## 1 Introduction

With the advancement of modern chemical synthesis platforms, the discovery of new molecules has become more efficient and data-driven models have become increasingly crucial for their generation and evaluation. Among these models, Quantitative Structure–Activity Relationships (QSARs) are a established approach to predict molecular properties that would otherwise require expensive and time-consuming experimentation to be obtained. However, the prediction accuracy of QSARs is limited, and their applicability can be hindered by the requirement for pre-defined structural features.[1] In recent years, neural networks (NNs) have been increasingly employed for molecular property prediction, offering a more flexible alternative to traditional QSARs.[2] Unlike QSARs, NNs can learn complex, data-driven representations of molecules tailored to specific tasks, without relying on pre-defined molecular features designated by experts.[3–6] However, the use of NNs in molecular modeling still has limitations, particularly in terms of expressiveness and transparency. The complexity of NN models can make it difficult to assess their robustness, out-of-domain applicability, and potential failure modes.

Incorporating uncertainty quantification (UQ) capabilities is essential to overcome the limitations of NNs in molecular modeling.[7] UQ refers to a set of mathematical techniques designed to quantify both aleatoric (or data) uncertainty and epistemic (or model) uncertainty. Aleatoric uncertainties arise from random noise in data observations or measurements and epistemic uncertainties arise from lack of knowledge, such as imbalanced data, insufficient data representation, or poor model architecture, as shown in Fig. 1a. While aleatoric uncertainty is typically considered to be irreducible, epistemic uncertainty can be reduced (e.g., by collecting additional training data in relevant regions of an experimental space).[7] By incorporating UQ techniques into NNs, valuable insights can be gained into the robustness and reliability of predictions, particularly in situations where data may be limited or noisy.[8] This is especially critical in molecular property predictions, where inaccurate predictions can have severe consequences. For example, a mispredicted toxicology profile of a drug candidate could result in costly clinical trials being terminated or can even lead to a product recall.[9] By quantifying uncertainties associated with molecular property predictions, UQ can help identify potential sources of error, enhance model transparency, and ultimately enable the development of more accurate and reliable models.[8] Moreover, data accessibility issues often necessitate the quantification of different uncertainty sources. Employing UQ can facilitate an understanding of the cost-benefit ratio of gathering new data, which is beneficial when it decreases epistemic uncertainties. However, if the prevailing uncertainties are aleatoric, which are inherently irreducible, the pursuit of new data may be unproductive.

[a]Department of Chemical and Biological Engineering, University of Wisconsin–Madison, 1415 Engineering Dr, Madison, WI 53706, USA. E-mail: sjiang87@wisc.edu

[b]Computing and Computational Sciences Directorate, Oak Ridge National Laboratory, P.O. Box 2008, Oak Ridge, TN 37831, USA

[c]Mathematics and Computer Science Division, Argonne National Laboratory, Lemont, IL 60439, USA

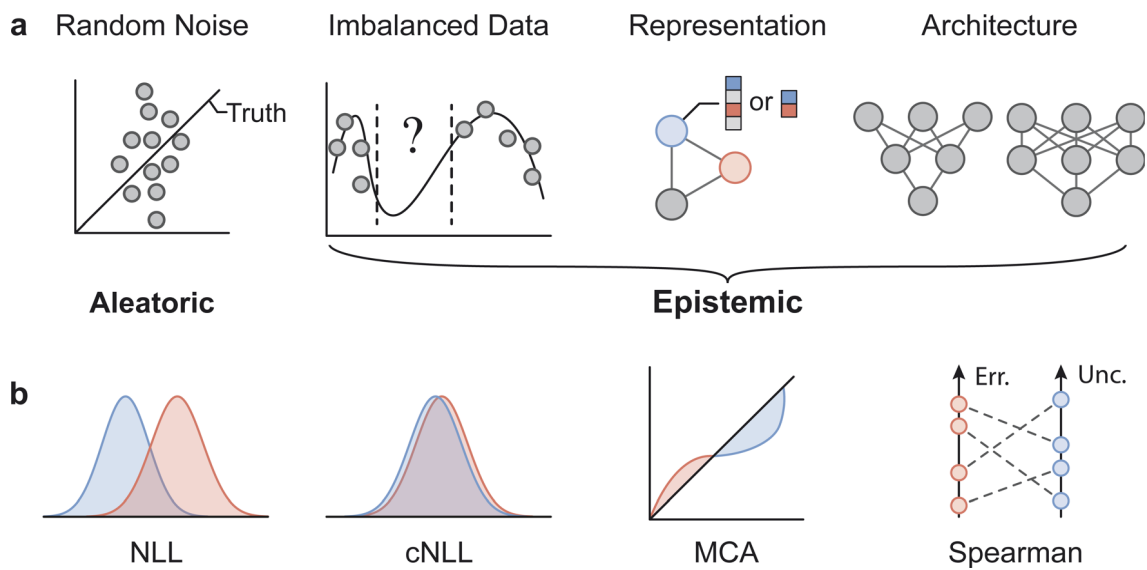

**Fig. 1** Sources of aleatoric and epistemic uncertainties and key metrics for quantifying uncertainty. (a) Aleatoric uncertainties arise from random noise in observations or measurements, while epistemic uncertainties arise from lack of information, such as imbalanced data, insufficient representation, and poor model architectures. (b) Common UQ metrics include negative log-likelihood (NLL), calibrated NLL, miscalibration area (MCA), and Spearman's rank correlation coefficient.

In recent years, significant effort has been devoted to developing UQ techniques for NN models in molecular prediction. Gawlikowski et al. have provided a comprehensive review of most UQ techniques,[10] while Hirschfeld et al. have applied multiple UQ techniques in molecular property prediction using graph NNs.[11] Various UQ techniques are often used, as summarized in Fig. 2. Mean-variance estimation[12] is a popular method for quantifying uncertainty in machine learning models; in this method, the mean ($\mu$) and variance ($\sigma^2$) of the model predictions are calculated over a set of inputs by assuming Gaussian noise associated with the prediction. The mean value represents the model best estimate of the predicted output, while the variance represents the uncertainty associated with that estimate.

Bayesian methods, another key paradigm of UQ, offer a probabilistic approach to modeling uncertainty in NNs. Monte Carlo dropout[7,13,14] is a widely used Bayesian method based on variational inference. This method involves randomly dropping out a proportion of the neurons during training and running multiple forward passes with different dropout masks to generate a distribution of predictions. Although Monte Carlo dropout has been shown to be effective in capturing model uncertainty and providing reliable uncertainty estimates,[7] it is limited by the training conditions of the original model, which can lead to highly correlated predictions across models.[15]

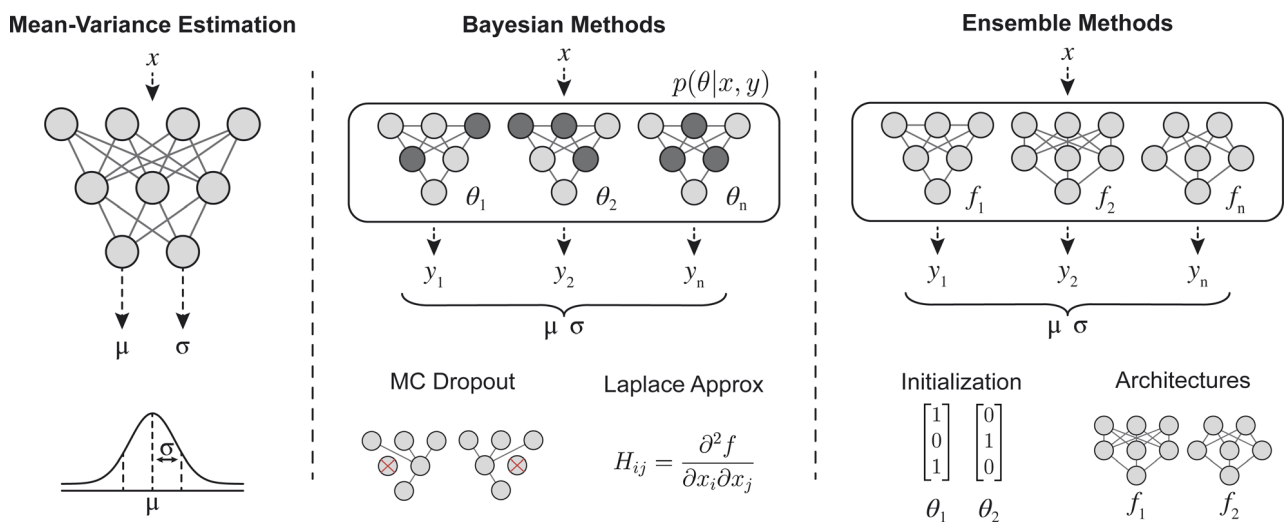

**Fig. 2** Various uncertainty quantification methods for neural networks.

The so-called Laplace approximation is another commonly used Bayesian UQ method that involves approximating the posterior distribution of NN parameters around a local mode of the loss surface using a multivariate Gaussian distribution. The estimation of the Hessian matrix of the NN is critical to the Laplace approximation, but this typically cannot be directly computed due to the massive number of parameters involved. Therefore, several techniques have been proposed, such as layer-wise Kronecker Factor approximation.[16,17] While the Laplace approximation requires only one NN model (thus being efficient),[18] it has some limitations. For example, it assumes that the posterior distribution is approximately Gaussian, which may not hold true far from the mode. Additionally, the Laplace approximation involves fidelity-complexity trade-offs and can be sensitive to the training conditions.

Ensemble methods are also widely used for UQ and involve the aggregation of predictions from multiple models, referred to as ensemble members, to derive a final prediction. This approach aims to enhance generalization by leveraging the complementary strengths of the individual models.[19] To maximize the diversity among the individual models, various techniques have been explored, including distinct random initialization of models,[20] Bagging and Boosting,[21] and ensembling of different network architectures.[22,23] Egele et al. recently demonstrated that the use of diverse network architectures can significantly increase the diversity within the ensemble, thereby leading to more accurate estimates of model uncertainty.[23] Moreover, they introduced a novel strategy for training multiple candidate models in parallel, which greatly reduces computational time. Ensemble methods have shown to be effective in achieving reliable uncertainty estimates and are easy to apply and parallelize, making them a practical choice for UQ applications.

Despite recent progress in UQ for NNs, several challenges remain to be addressed. One of the most significant challenges is the need to separate aleatoric and epistemic uncertainty. This separation is crucial, as it enables us to understand the sources of uncertainty in the model and provides insight into how we can improve it. For instance, quantifying epistemic uncertainty can inform the need to collect more data or improve the model architecture, such as incorporating additional features or using a more complex NN. Moreover, the accuracy of the model prediction is fundamental to the quality of UQ. Without an accurate model prediction, the UQ output will be unreliable and potentially misleading.

To address these challenges, this paper proposes an approach, which we call AutoGNNUQ, for constructing a diverse ensemble of GNN models for molecular property prediction by adapting the AutoDEUQ method.[23] AutoGNNUQ employs an aging evolution (AE)[24] approach to search for network architectures, with each model trained to minimize the negative log-likelihood to capture aleatoric uncertainty. The approach then selects a set of models from the search to construct the ensembles and model epistemic uncertainty without sacrificing the quality of aleatoric uncertainty. The method achieves high prediction accuracy and UQ performance on several benchmark datasets, outperforming existing algorithms. Moreover, the decomposition of aleatoric and epistemic uncertainty provides insights into possible areas for reducing uncertainty. The proposed approach has the potential to significantly enhance UQ techniques for GNN models in molecular prediction, leading to more reliable and efficient active learning and experimental design. We provide extensive benchmark results against established UQ paradigms and datasets to demonstrate the benefits of the proposed approach. We provide all data and code needed for implementing our approach and for reproducing the benchmark results.

## 2 Methods

The AutoGNNUQ workflow, illustrated in Fig. 3, consists of three key steps which we will detail in this section. First, molecular data are represented as graphs with atomic and bond features. Second, a neural architecture search (NAS) algorithm is employed to identify high-performing GNN models for UQ. Third, a diverse ensemble of these high-performing models is assembled, resulting in accurate molecular property prediction, high UQ performance, and decomposition of total uncertainty into aleatoric and epistemic uncertainties.

### 2.1 Dataset and representation

We evaluated the efficacy of the proposed AutoGNNUQ approach for molecular property prediction and UQ using benchmark datasets from MoleculeNet[25] for Lipo, ESOL, FreeSolv, QM7 and QM9. These datasets were also employed in the previous UQ benchmark study.[11] The Lipo dataset[26] contains 4306 compounds with measured octanol–water partition coefficients ($\log P$). It aims to predict the lipophilicity of compounds, which is an essential property in drug discovery as it influences drug absorption, distribution, and metabolism within the human body. The FreeSolv dataset[27] comprises 643 small molecules with the hydration free energy ($\Delta G_{hyd}$) as a thermodynamic quantity that characterizes the interaction between a solute and solvent. The ESOL dataset[28] includes 1128 compounds with measured solubility in water ($S_{water}$) ranging from 0.001 to 10 000 mg $L^{-1}$. The QM7 dataset[29,30] contains atomization energies ($\Delta H_{atom}^{\circ}$) of 7211 organic molecules with up to seven heavy atoms (i.e., C, N, O, and S). Finally, the QM9 dataset[31] provides twelve properties for a subset of 133 865 molecules from the GDB-17 database,[32] each containing up to nine heavy atoms (C, N, O, and F). These properties include dipole moment ($\mu$), isotropic polarizability ($\alpha$), highest occupied molecular orbital energy ($\varepsilon_{HOMO}$), lowest unoccupied molecular orbital energy ($\varepsilon_{LUMO}$), gap between $\varepsilon_{HOMO}$ and $\varepsilon_{LUMO}$ ($\Delta\varepsilon$), electronic spatial extent ($<R^2>$), zero point vibrational energy (ZPVE), internal energy at 0 K ($U_0$), room temperature internal energy ($U$), enthalpy ($H$), free energy ($G$), and heat capacity ($c_v$).

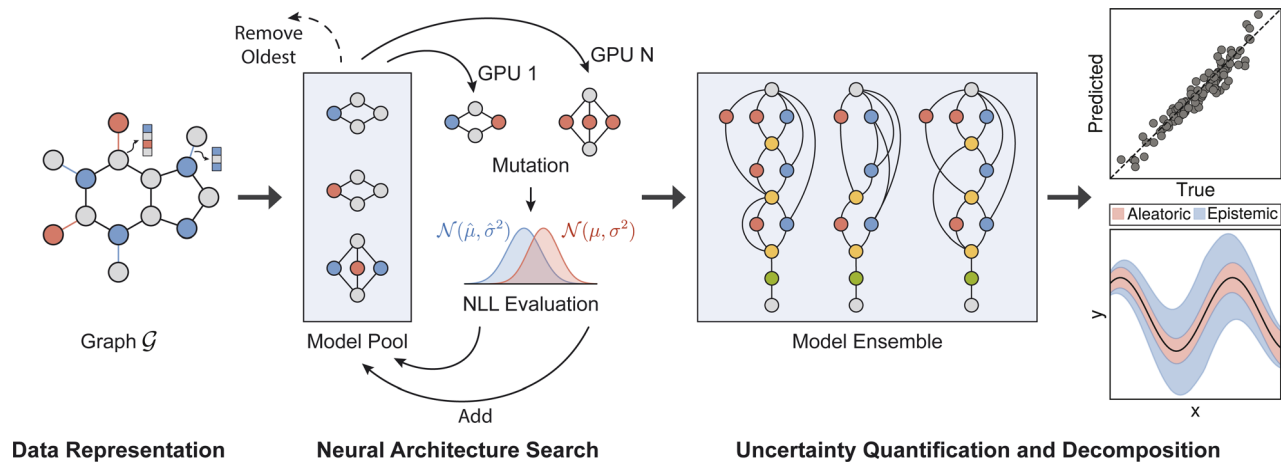

**Fig. 3** Workflow of AutoGNNUQ for uncertainty quantification in molecular property prediction. A molecular structure is represented as a graph with node features representing atoms and edge features representing bonds. Graph neural networks are used for uncertainty quantification (UQ). To optimize UQ performance, a neural architecture search algorithm using aging algorithms continuously mutates the architecture and evaluates UQ performance using the negative log-likelihood loss function. The resulting high-performing models are added to a model pool, and the oldest models are removed. From this pool, an ensemble of models is selected to achieve high molecular property prediction accuracy and UQ performance as well as to decompose aleatoric and epistemic uncertainties.

The properties predicted and their respective units for each dataset are detailed in Table 1.

For consistency with prior research, we utilized identical graph representations of molecules where atoms serve as nodes and bonds as edges. Additionally, we adopted the same features used in the benchmark investigation,[11] which consist of 133 atomic features and 14 bond features. The atomic features encompass several characteristics such as atomic number, degree, formal charge, chirality, number of hydrogens, and types of hybridization. The bond features, on the other hand, capture properties such as bond type, chirality, whether it is conjugated, and whether it is in a ring. It is important to note that although the atomic feature vector is extensive, the first 100 entries primarily consist of the one-hot encoding of atomic numbers. We further examined how representation affects prediction accuracy and UQ performance by using the simplest set of features, with only atomic numbers as atomic features and bond orders as bond features.

## 2.2 Neural architecture search and ensemble construction

The AutoGNNUQ NAS process consists of two main components, which are detailed below. First, a search space is established that comprises a range of feasible architectures using a message passing neural network (MPNN), a type of GNN, for extracting molecular features. Second, a search method is employed to explore this search space and identify the high-performing MPNN architecture for UQ. To generate a catalog of NNs, we execute AutoGNNUQ and store all models from the runs. To construct an ensemble $\mathcal{E}$ of models from the model pool $\mathcal{C}$, we use a top-K approach, selecting the top K architectures with the lowest validation loss (*i.e.*, negative log-likelihood). Specifically, we set K to 10 to ensure sufficient diversity and model representation in the ensemble.

**Table 1** Property prediction results comparison. Mean values are reported with standard deviations in parentheses. The best result is bold, and the second-best result is in italic

| Dataset | Property | Unit | AutoGNNUQ | AutoGNNUQ-simple | MC dropout | Random ensemble | Benchmark |
|---|---|---|---|---|---|---|---|
| Lipo | $\log P$ | $\log D$ | **0.64 (0.02)** | 0.89 (0.15) | *0.67 (0.03)* | 0.75 (0.02) | 0.73 (0.11) |
| ESOL | $S_{\text{water}}$ | log mol L$^{-1}$ | *0.74 (0.06)* | 0.82 (0.10) | 0.76 (0.06) | 0.80 (0.04) | **0.58 (0.03)** |
| FreeSolv | $\Delta G_{\text{hyd}}$ | kcal mol$^{-1}$ | *1.32 (0.29)* | 1.62 (0.35) | 1.33 (0.30) | 1.43 (0.12) | **1.15 (0.12)** |
| QM7 | $\Delta H_{\text{atom}}$ | kcal mol$^{-1}$ | *47.5 (2.1)* | **46.3 (1.5)** | 49.9 (3.9) | 58.6 (6.6) | 77.9 (2.1) |
| QM9 | $\mu$ | D | *0.585 (0.038)* | 0.700 (0.079) | 0.618 (0.040) | 0.853 (0.021) | **0.358** |
| | $\alpha$ | $a_0^3$ | **0.329 (0.025)** | *0.438 (0.074)* | 0.626 (0.041) | 1.660 (0.393) | 0.890 |
| | $\varepsilon_{\text{HOMO}}$ | eV | **0.109 (0.016)** | 0.146 (0.023) | *0.138 (0.009)* | 0.246 (0.024) | 0.147 |
| | $\varepsilon_{\text{LUMO}}$ | eV | **0.107 (0.011)** | 0.154 (0.032) | *0.145 (0.008)* | 0.383 (0.069) | 0.170 |
| | $\Delta \varepsilon$ | eV | **0.149 (0.015)** | *0.210 (0.041)* | 0.192 (0.012) | 0.441 (0.068) | 0.223 |
| | $<R^2>$ | $a_0^2$ | **28.1 (2.0)** | 34.8 (4.5) | 37.1 (1.9) | 83.2 (15.1) | *28.5* |
| | ZPVE | eV | **0.00509 (0.00048)** | *0.00683 (0.00099)* | 0.051 (0.013) | 0.186 (0.060) | 0.0588 |
| | $c_v$ | cal mol$^{-1}$ K | **0.156 (0.016)** | *0.216 (0.037)* | 0.349 (0.036) | 0.946 (0.256) | 0.42 |
| | $U_0$ | kcal mol$^{-1}$ | **0.0146 (0.0049)** | *0.0207 (0.0078)* | 1.52 (0.16) | 4.26 (2.29) | 2.05 |
| | $U$ | kcal mol$^{-1}$ | **0.0138 (0.0053)** | *0.0205 (0.0072)* | 1.52 (0.16) | 4.26 (2.29) | 2.00 |
| | $H$ | kcal mol$^{-1}$ | **0.0157 (0.0067)** | *0.0202 (0.0078)* | 1.51 (0.17) | 4.26 (2.29) | 2.02 |
| | $G$ | kcal mol$^{-1}$ | **0.0143 (0.0060)** | *0.0212 (0.0067)* | 1.51 (0.17) | 4.26 (2.29) | 2.02 |

**2.2.1 Search space.** We have defined the AutoGNNUQ search space as a directed acyclic graph, illustrated in Fig. 4. The search space has fixed input and output nodes denoted by $\mathcal{I}$ and $\mathcal{O}$, respectively. All other nodes, denoted by $\mathcal{N}$, represent intermediate nodes and contain a set of feasible operations. These intermediate nodes can be categorized into two types: constant nodes that contain a single operation, and variable nodes that contain multiple operations. For each variable node, an index is assigned to each operation. An architecture in the search space can be defined using a vector $\mathbf{p} \in \mathbb{Z}^n$, where $n$ is the number of variable nodes. Each entry $\mathbf{p}_i$ represents an index chosen from a set of feasible index values for the variable node $i$. The AutoGNNUQ search space is composed of MPNN, skip-connection, and gather variable nodes, which are further elaborated below.

1. *Input node*: the input node consists of several components, including node features, edge features, edge pairs, and node masks. To construct the feature matrices for a given dataset of molecules, we consider the maximum number of nodes and edges as $N$ and $E$, respectively. We pad the node feature matrix with zeros to generate $\mathbf{H} \in \mathbb{R}^{N \times F_n}$, and the edge feature matrix with zeros to generate $\mathbf{E} \in \mathbb{R}^{E \times F_e}$, where $F_n$ and $F_e$ represent the number of node features and edge features, respectively. Additionally, we employ an edge pair matrix $\mathbf{P} \in \mathbb{Z}^{E \times 2}$, where each row denotes the indices of two nodes connected by an edge. Given that molecules can have different numbers of nodes, we use a node mask vector $\mathbf{m} \in \mathbb{Z}^N$ to exclude non-existent node features that occur from zero padding. A present node is denoted by $\mathbf{m}_i = 1$, while a non-existent node is indicated by $\mathbf{m}_i = 0$.

2. *MPNN node*: the MPNN node updates the hidden features of each node for $T$ time steps using a message function $M_t$ and an update function $U_t$. These functions are defined as follows:

$$\mathbf{m}_v^{t+1} = \text{Agg}_{w \in \mathcal{N}(v)} M_t(\mathbf{h}_v^t, \mathbf{h}_w^t, \mathbf{e}_{vw}) \quad (1a)$$

$$\mathbf{h}_v^{t+1} = U_t(\mathbf{h}_v^t, \mathbf{m}_v^{t+1}). \quad (1b)$$

To update the hidden feature of a node $v$ at step $t$, the message function $M_t$ takes as inputs the node $v$ feature $\mathbf{h}_v^t$, the neighboring node features $\mathbf{h}_w^t$ for $w \in \mathcal{N}(v)$, and the edge feature $\mathbf{e}_{vw}$ between node $v$ and $w$. The output of the message function $M_t$ is a list of message vectors from neighboring nodes, which are collected and used by the aggregate function Agg to generate the intermediate hidden feature $\mathbf{w}_v^{t+1}$. The aggregate function can be one of mean, summation or max pooling. At each step $t$, the update function $U_t$ merges the node feature $\mathbf{h}_v^t$ with the intermediate hidden feature $\mathbf{m}_v^{t+1}$ to generate the new hidden feature $\mathbf{h}_v^{t+1}$ for the next step $t + 1$. The message function $M_t$ and update function $U_t$ are described in detail as follows:

$$M_t(\mathbf{h}_v^t, \mathbf{h}_w^t, \mathbf{e}_{vw}) = \alpha_{vw} \text{MLP}(\mathbf{e}_{vw}) \mathbf{h}_w^t \quad (2a)$$

$$U_t(\mathbf{h}_v^t, \mathbf{m}_v^{t+1}) = \begin{cases} \text{GRU}(\mathbf{h}_v^t, \mathbf{m}_v^{t+1}) \\ \text{MLP}(\mathbf{h}_v^t, \mathbf{m}_v^{t+1}) \end{cases}. \quad (2b)$$

The message function involves a multi-layer perceptron (MLP), also known as an edge network, to manage the edge feature $\mathbf{e}_{vw}$. After being processed by the MLP, the edge feature is multiplied by $\mathbf{h}_w^t$ to produce a message from node $w$ to $v$. In this situation, the processed edge feature $\text{MLP}(\mathbf{e}_{vw})$ can be considered as a weight for $\mathbf{h}_w^t$. Building on the concept of node attention, we introduce an attention coefficient $\alpha_{vw}$ to adjust the weight of $\mathbf{h}_w^t$. This coefficient is determined by a function of both $\mathbf{h}_v^t$ and $\mathbf{h}_w^t$. The update function $U_t$ can either be a gated recurrent unit (GRU) or a multi-layer perceptron (MLP). To provide additional clarity regarding the design of the MPNN node, we can break it down into the following five operational categories.

(a) *Hidden dimension*: after $T$ iterations, the MPNN node transforms the input node feature into a $d$-dimensional vector. The selection of the hidden dimension $d$ plays an important role in the prediction. In order to enhance generalization and reduce the number of parameters, we select the set of state dimensions to $\{8, 16, 32, 64\}$.

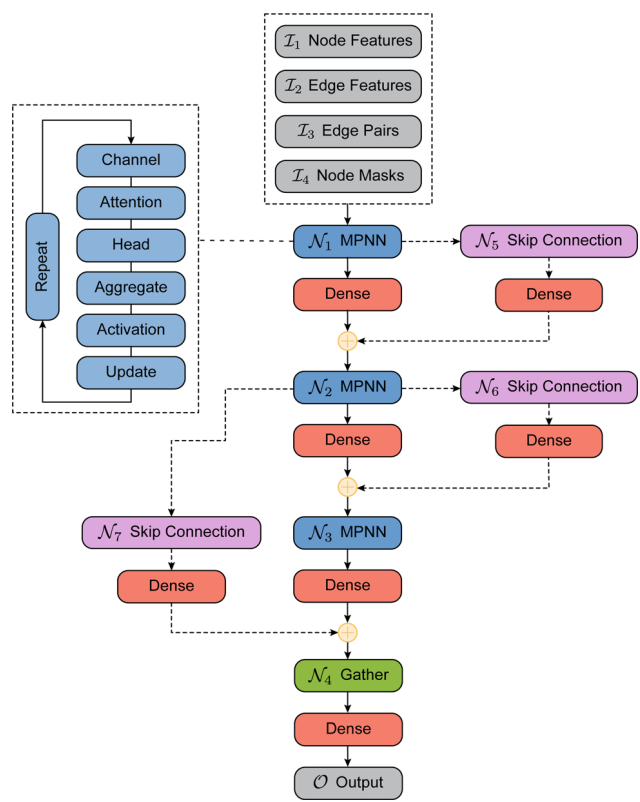

**Fig. 4** Example AutoGNNUQ search space with three MPNN variable nodes in blue ($\mathcal{N}_1$, $\mathcal{N}_2$, and $\mathcal{N}_3$), skip-connection variable nodes in pink ($\mathcal{N}_5$, $\mathcal{N}_6$, and $\mathcal{N}_7$), and a gather variable node in green ($\mathcal{N}_4$). Dotted lines represent possible skip connections. The MPNN takes as input node features ($\mathcal{I}_1$), edge features ($\mathcal{I}_2$), edge pairs ($\mathcal{I}_3$), node masks ($\mathcal{I}_4$), and consists of two constant dense nodes with 32 hidden units before outputting a single node ($\mathcal{O}$). The MPNN node includes several variables such as the number of channels, attention mechanisms, number of attention heads, aggregation methods, activation functions, update functions, and number of repetitions.

(b) *Attention function*: while conventional MPNNs rely on the edge feature to dictate the weight of information propagation between nodes, the attention mechanism enables prioritization of the most pertinent neighboring nodes, thereby improving the process of information aggregation. Drawing upon our prior research,[33] we employ a range of attention functions including Constant, GAT, SYM-GAT, COS, Linear, and Gen-linear (see details in ESI).†

(c) *Attention head*: the use of multi-head attention can be advantageous in ensuring stable learning.[34] We select the number of heads from the set of $\{1, 2, 3\}$.

(d) *Aggregate function*: the selection of an appropriate aggregation function is crucial in capturing neighborhood structures and extracting node representation.[35] We choose our aggregation functions from the set of {mean, summation, max-pooling}.

(e) *Activation function*: as per our prior research,[33] we consider a range of activation functions {Sigmoid, Tanh, ReLU, Linear, Softplus, LeakyReLU, ReLU6, and ELU}.

(f) *Update function*: combining node features $\mathbf{h}_v^t$ with intermediate hidden features $\mathbf{m}_v^{t+1}$ and propagating them using an update function enables the generation of new features, denoted as $\mathbf{h}_v^{t+1}$. Our update functions are chosen from the set {GRU, MLP}.

3. *Skip-connection node*: the skip-connection node is a type of variable node that enables a connection between nodes $\mathcal{N}_{i-1}$, $\mathcal{N}_i$, and $\mathcal{N}_{i+1}$ in a sequence. The purpose of the skip-connection is to add or skip a node within a sequence. This operation involves two possible actions: identity for skip-connection or empty for no skip-connection. In a skip-connection operation, the tensor output from $\mathcal{N}_{i-1}$ is processed by a dense layer, which ensures the incoming tensor is projected to the appropriate shape for summation. The summation operator is then applied to add the output from $\mathcal{N}_{i-1}$ and $\mathcal{N}_i$, producing a result that is passed to $\mathcal{N}_{i+1}$. The skip-connection can be applied to any length of node sequences, but in this study, it is limited to a maximum of three nodes to restrict complexity. For instance, $\mathcal{N}_{i-1}$ may be added to $\mathcal{N}_{i+2}$ and passed to $\mathcal{N}_{i+3}$.

4. *Gather variable node*: the gather node consists of eleven operations categorized into five different types. The input to the gather node is the node feature matrix $\mathbf{H} \in \mathbb{R}^{N \times F}$, where $N$ represents the number of nodes and $F$ represents the number of hidden features. The graph operations used in the AutoGN-NUQs can be segregated into five categories based on the gather operations provided in the Spektral GNN package.[36]

(a) *Global pool*: aggregate node features through the computation of the sum, mean, or maximum. The output has a shape represented by $\mathbb{R}^N$.

(b) *Global gather*: compute the sum, mean, or maximum of a feature for all the nodes. The output has a shape of $\mathbb{R}^F$.

(c) *Global attention pool*: calculate the output $\mathbf{H}_{out} \in \mathbb{R}^{F'}$ as

$$\mathbf{H}_{out} = \sum_{i=1}^{N} (\sigma(\mathbf{HW}_1 + \mathbf{b}_1) \odot (\mathbf{HW}_2 + \mathbf{b}_2))_i$$

Here, $\sigma$ represents the sigmoid activation function, and $\mathbf{W}_1$ and $\mathbf{W}_2$ represent trainable weights, while $\mathbf{b}_1$ and $\mathbf{b}_2$ represent biases. The output dimension $F'$ can be selected from the set $\{16, 32, 64\}$.

(d) *Global attention sum pool*: aggregate a graph by learning attention coefficients to sum node features. The operation can be defined as follows: $\mathbf{H}_{out} = \sum_{i=1}^{N} \alpha_i \mathbf{H}_i$, where $\alpha$ is defined as softmax$(\mathbf{Ha})$, and $\mathbf{a}$ is a trainable vector. The softmax activation is applied across all the nodes.

(e) *Flatten*: flatten $\mathbf{H}$ into a 1D vector.

The total number of possible architectures in the search space is $12\,259\,638\,116\,352$ ($\approx 1.2 \times 10^{13}$). This highlights the rich model space that the proposed AutoGNNUQ approach has available.

**2.2.2 Search method.** In order to discover high-performing neural architectures from the search space, we use the aging evolution (AE) method,[24] which is an asynchronous search technique that is available in the DeepHyper package.[37] The AE algorithm starts with a population of $P$ random architectures, evaluates them, and records the validation loss of each one. After the initialization, the algorithm samples $N$ random architectures uniformly from the population, and the architecture with the lowest validation loss in the sample is chosen as the parent. A mutation is then applied to the parent model by selecting a random and different operation for a random variable node, while all other variable nodes remain fixed. This creates a new child architecture that is trained, and its validation loss is recorded. The child architecture is added to the population by replacing the oldest architecture in the population. Over multiple cycles, the algorithm retains architectures with lower validation loss *via* repeated sampling and mutation. AE is highly scalable as it can leverage multiple compute nodes to evaluate architectures in parallel, resulting in faster convergence to high-performing architectures. AE has been shown to outperform reinforcement learning methods for NAS due to its minimal algorithmic overhead and synchronization.[24,38] In contrast to non-aging evolution (NAE) methods, such as standard tournament selection, where the best-performing models can dominate the population and reduce exploration, AE promotes greater diversity by frequently renewing the population. In AE, models have a short lifespan, requiring architectures to perform well consistently in retraining to persist through generations. This process effectively focuses on robust architectures rather than those that may have performed well initially by chance. The frequent renewal leads to increased exploration and a more diverse set of architectures.[24]

**2.2.3 Search process.** We adopt the same approach for splitting the data as the benchmark study[11] to ensure consistency and comparability. We use the identical 8 random seeds as used in ref. 11 for randomly splitting the data into training, validation, and testing sets. The split ratio is 5 : 2 : 3 for the Lipo, ESOL, FreeSolv, and QM7 datasets. For QM9, the random split ratio is 8 : 1 : 1 to match the MoleculeNet benchmark.[25] The training set is utilized to optimize the model parameters, while the validation set guides the NAS. Finally, the performance of the model is evaluated using the testing set. During the search process, we imposed a time limit on the training for each architecture, limiting all datasets to 30 epochs. For Lipo, the batch size was set to 128, while for the other datasets it was set to 512. We utilized the Adam[39] optimizer with a learning rate of 0.001 and performed 1000 architecture searches. Once the search was complete, we performed post-training by selecting 10 architectures with the lowest validation loss and training them from scratch for 1000 epochs to form the ensemble $\mathcal{E}$.

### 2.3 Uncertainty quantification and decomposition

The dataset, denoted by $\mathcal{D}$, comprises input graphs $\mathbf{x} \in \mathcal{X}$ and their corresponding outputs $y \in \mathcal{Y}$, where $\mathcal{X}$ and $\mathcal{Y}$ represent the input and output spaces, respectively. We focus on regression problems in which the output is a scalar or vector of real values. Our objective is to model the probabilistic predictive distribution $p(y|\mathbf{x})$ using a parameterized distribution $p_\theta(y|\mathbf{x})$, which estimates aleatoric uncertainty. To capture epistemic uncertainty, we use an ensemble of neural networks denoted by $p_\mathcal{E}(y|\mathbf{x})$, where $\mathcal{E}$ represents the set of all models in the ensemble. The sample space for $\theta$ is defined as $\Theta$, which represents the space of all possible values for the parameters in the parameterized distribution. By combining both aleatoric and epistemic uncertainties, we aim to improve the overall predictive performance and UQ in molecular property prediction tasks.

#### 2.3.1 Aleatoric uncertainty.
To model aleatoric uncertainty, we use the quantiles of the probability distribution $p_\theta$, where $\theta$ is partitioned into architecture decision variables $\theta_a$ (representing the network topology parameters) and model weights $\theta_w$. We assume a Gaussian distribution for $p_\theta$, such that $p_\theta \sim \mathcal{N}(\mu_\theta, \sigma_\theta^2)$, and measure aleatoric uncertainty using variance, consistent with previous works.[20,23] The NN is trained to output both the mean $\mu_\theta$ and the variance $\sigma_\theta^2$ (which is essentially mean-variance estimation). To obtain the optimal choice of $\theta_w$ given a fixed $\theta_a$, we seek to maximize the likelihood of the real data $\mathcal{D}$. This is achieved by minimizing the negative log-likelihood loss function,[20] which is given by:

$$\mathscr{L}(\mathcal{D}; \theta) = -\log p_\theta$$

$$= \frac{1}{2|\mathcal{D}|} \sum_{\mathbf{x}, y \in \mathcal{D}} \left( \log(2\pi) + \log\left(\sigma_\theta(\mathbf{x})^2\right) + \frac{(\mu_\theta(\mathbf{x}) - y)^2}{\sigma_\theta(\mathbf{x})^2} \right). \quad (3)$$

#### 2.3.2 Epistemic uncertainty.
To account for epistemic uncertainty, we employ ensembles consisting of multiple neural networks (NNs).[20] Our approach involves generating a collection of NNs, denoted by $\mathcal{C} = \theta_i, i = 1, 2, ..., c$, where each $\theta$ is a combination of architecture decision variables and model weights. We then select $K$ models from this collection to form the ensemble, where $\mathcal{E} = \theta_i, i = 1, 2, ..., K$ and $K$ denotes the ensemble size.

To describe the probability of $\theta \in \mathcal{E}$, we define a probability measure $p : \Theta \rightarrow [0, 1] \in \mathbb{R}$ such that $p(\theta)$ represents the probability of $\theta$ being present in the ensemble. To obtain the overall probability density function of the ensemble and effectively model epistemic uncertainty, we use a mixture distribution, which is a weighted average of the probability density functions of all the members in the ensemble. Specifically, we have:

$$p_\mathcal{E} = \mathbb{E}_{\theta \sim p(\mathcal{E})} p_\theta, \quad (4)$$

where $p_\theta$ represents the parameterized distribution that estimates the probabilistic predictive distribution $p(y|\mathbf{x})$ for a given $\theta$, and $p(\mathcal{E})$ refers to a probability distribution over the ensemble. This approach allows us to effectively model epistemic uncertainty by capturing the diversity of predictions made by different NNs in the ensemble.

#### 2.3.3 Uncertainty decomposition.
By the law of total variance,[23] we have that

$$\mu_\mathcal{E} := \mathbb{E}_{\theta \sim p(\mathcal{E})}[\mu_\theta] \quad (5a)$$

$$\sigma_\mathcal{E}^2 := \mathbb{V}_{\theta \sim p(\mathcal{E})}[p_\mathcal{E}]$$

$$= \underbrace{\mathbb{E}_{\theta \sim p(\mathcal{E})}\left[\sigma_\theta^2\right]}_{\text{Aleatoric Uncertainty}} + \underbrace{\mathbb{V}_{\theta \sim p(\mathcal{E})}[\mu_\theta]}_{\text{Epistemic Uncertainty}}, \quad (5b)$$

where $\mathbb{E}$ refers to the expected value and $\mathbb{V}$ refers to the variance.

Eqn (5b) provides a formal decomposition of the overall uncertainty of the ensemble into its individual components such that

- $\mathbb{E}_{\theta \sim p(\mathcal{E})}[\sigma_\theta^2]$ marginalizes the effect of $\theta$ and characterizes the aleatoric uncertainty.
- $\mathbb{V}_{\theta \sim p(\mathcal{E})}[\mu_\theta]$ captures the spread of the prediction across different models and ignores the noise present in the data, therefore characterizing the epistemic uncertainty.

If we assume that $p(\mathcal{E})$ is uniform in the mixture distribution, meaning that the weights are equal, we can calculate the two values in eqn (5b) using empirical mean and variance estimates,

$$\mu_\mathcal{E} = \frac{1}{K} \sum_{\theta \in \mathcal{E}} \mu_\theta \quad (6a)$$

$$\sigma_\mathcal{E}^2 = \underbrace{\frac{1}{K} \sum_{\theta \in \mathcal{E}} \sigma_\theta^2}_{\text{Aleatoric Uncertainty}} + \underbrace{\frac{1}{K - 1} \sum_{\theta \in \mathcal{E}} (\mu_\theta - \mu_\mathcal{E})^2}_{\text{Epistemic Uncertainty}} \quad (6b)$$

where $K$ is the size of the ensemble.

In this context, we have established that the overall uncertainty $\sigma^2$ is the sum of two distinct sources, namely the aleatoric and epistemic uncertainty. The aleatoric uncertainty is determined by the mean value of the predictive variance of each model in the ensemble, whereas the epistemic uncertainty is determined by the predictive variance of the mean value of each model in the ensemble. We note that recent advancements[40,41] in evidential UQ can also track both aleatoric and epistemic uncertainties by parameterizing the posterior distribution as a Normal Inverse-Gamma (NIG) distribution and inferring these uncertainties from the NIG distribution parameters. A comparison of evidential UQ and AutoGNNUQ in terms of the confidence curve is shown in ESI Table S6.† Incorporating the evidential model as candidate models in AutoGNNUQ is an important future step.

#### 2.3.4 Monte Carlo dropout.
Monte Carlo (MC) dropout involves training a GNN with dropout before each layer and maintaining dropout during testing to generate $N$ outputs using random dropout masks. Each mask is a sample from the approximate posterior distribution. The dropout ratio is set to 0.1, and $N$ is fixed at 10, aligning with the ensemble size in AutoGNNUQ. MC dropout uses the model architecture from AutoGNNUQ with the lowest validation loss for a fair comparison. Mean, aleatoric, and epistemic uncertainties are calculated using the same approach as in AutoGNNUQ.

### 2.4 Out-of-distribution performance

Ensuring that UQ methods are effective for out-of-distribution (OOD) molecules, which significantly differ from training data, is essential for optimally selecting new molecules for characterization. Inspired by a prior OOD study,[42] we assessed the AutoGNNUQ model trained on the QM9 dataset and evaluated its prediction accuracy and UQ performance on the PC9 dataset.[43] Specifically, HOMO and LUMO properties in a subset of PC9 that does not overlap with QM9 are analyzed. While both datasets limit molecules to having up to nine heavy atoms, PC9 uniquely includes species with multiplicities greater than 1, totaling 5351 of the 78 172 molecules. This subset includes 4468 radicals with multiplicities of 2 and 883 triplets. We computed MACCS fingerprints for each molecule,[44] which encode chemical functional groups and substructures, and examined the distributional differences in their first principal component using principal component analysis (PCA). Additional details on physical and chemical differences, such as variations in bond distances and functional groups between QM9 and PC9, are available in the referenced study.[43] The property prediction and UQ on PC9 molecules are based on an ensemble of 80 models, consisting of the ten best models from each of the eight random seeds of AutoGNNUQ trained on QM9.

### 2.5 Evaluation metrics

To evaluate the performance of different UQ strategies, we employ the following metrics outlined in the benchmark study.[11]

**2.5.1 Negative log-likelihood (NLL).** One of the primary metrics used in our benchmark study is the negative log-likelihood (NLL), which quantifies the likelihood of the observed errors under the assumption that they follow a normal distribution with variances derived from the UQ estimates, denoted as $\sigma(\mathbf{x})^2$.[20] The NLL is calculated by taking the average of the negative logarithm of the likelihood function across all test molecules, and can be expressed as:

$$\text{NLL}(\mathcal{D}_{\text{test}}) = \frac{1}{2|\mathcal{D}_{\text{test}}|} \sum_{\mathbf{x}, y \in \mathcal{D}_{\text{test}}} \left( \log(2\pi) + \log\left(\sigma(\mathbf{x})^2\right) + \frac{(\mu(\mathbf{x}) - y)^2}{\sigma(\mathbf{x})^2} \right) \quad (7)$$

where $\mu(\mathbf{x})$ is the model prediction and $\sigma(\mathbf{x})^2$ is the UQ estimate for molecule $\mathbf{x}$.

**2.5.2 Calibrated NLL.** The uncalibrated NLL metric solely captures the difference between the predicted and true probability distributions. For reliable estimation of uncertainty, the predicted probabilities must be calibrated to match the true probabilities. Instead of considering the true variance as $\sigma(\mathbf{x})^2$, it is instead assumed to be linearly correlated as $a\sigma(\mathbf{x})^2 + b$.[45]

The calibrated NLL (cNLL) is determined for each dataset and method by minimizing the NLL in the validation set through the selection of scalars $a$ and $b$.

$$\text{cNLL} = \frac{1}{2|\mathcal{D}_{\text{test}}|}$$

$$\frac{1}{2|\mathcal{D}_{\text{test}}|} \sum_{\mathbf{x}, y \in \mathcal{D}_{\text{test}}} \left( \log(2\pi) + \log\left(a_*\sigma(\mathbf{x})^2 + b_*\right) + \frac{(\mu(\mathbf{x}) - y)^2}{a_*\sigma(\mathbf{x})^2 + b_*} \right) \quad (8a)$$

$$(a_*, b_*) = \operatorname*{argmin}_{\mathbf{x}, y \in \mathcal{D}_{\text{val}}} \left( \log(2\pi) + \log\left(a\sigma(\mathbf{x})^2 + b\right) + \frac{(\mu(\mathbf{x}) - y)^2}{a\sigma(\mathbf{x})^2 + b} \right) \quad (8b)$$

**2.5.3 Spearman's rank correlation coefficient.** Spearman's (rank correlation) coefficient is a useful tool in UQ for assessing the correlation between predicted uncertainties and actual errors. In any uncertainty estimator, we expect that predictions with lower uncertainty will be associated with lower true error. Specifically, if we have a model and two molecules $\mathbf{a}$ and $\mathbf{b}$ for which $\sigma_\mathbf{a}^2 < \sigma_\mathbf{b}^2$, we expect that $\mu_\mathbf{a}$ will be more accurate on average than $\mu_\mathbf{b}$.

To measure the correlation between an absolute error vector $\mathbf{m}$ and a predicted uncertainty vector $\mathbf{n}$, we define rank vectors $r_\mathbf{m}$ and $r_\mathbf{n}$, which assign integer ranks to each value in ascending order. The correlation coefficient is then calculated based on the covariance (cov) and standard deviations $\sigma$ of the rank vectors as

$$\rho(\mathbf{m}, \mathbf{n}) = \frac{\text{cov}(r_\mathbf{m}, r_\mathbf{n})}{\sigma(r_\mathbf{m})\sigma(r_\mathbf{n})}. \quad (9)$$

If $\mathbf{m}$ and $\mathbf{n}$ have the same ranking, $\rho(\mathbf{m}, \mathbf{n})$ is 1, and if $\mathbf{m}$ and $\mathbf{n}$ have opposite rankings, $\rho(\mathbf{m}, \mathbf{n})$ is $-1$. However, it is important to note that we do not expect a perfect correlation of $\rho = 1$ since we assume that the errors will follow a normal distribution. Thus, it is possible for the model to occasionally produce a result with low error, even if it has high uncertainty.

**2.5.4 Confidence curve.** A different ranking-based method for evaluating uncertainty, other than Spearman's coefficient, is the confidence curve. This method assesses how the error (measured as MAE or RMSE) varies when data points with the highest uncertainty are removed from the test dataset. A meaningful uncertainty measure should result in a lower error for a subset of high-confidence predictions. The confidence curve illustrates this by showing how the error changes as a function of confidence percentile. A steeper confidence curve indicates a more effective uncertainty estimate, as it signifies a faster decrease in error with the removal of high-uncertainty data. The ideal scenario is represented by the oracle confidence curve, where data points are ordered perfectly by their true error.[46] To quantify the difference between the oracle and the observed confidence curves, the area under the confidence-oracle error (AUCO) is calculated.[47] A smaller AUCO value indicates better performance, as it reflect a closer approximation to the ideal uncertainty ordering.

**2.5.5 Confidence-based calibration.** Confidence-based calibration[48,49] conceptualizes each prediction and its associated uncertainty as the mean and variance of a Gaussian distribution, denoted by $p(\mathbf{y}|\mathbf{x}) = \mathcal{N}(\mu(\mathbf{x}), \sigma(\mathbf{x})^2$. For a model to be deemed well-calibrated, a proportion of $x\%$ predictions should fall within the $x\%$ confidence interval. Specifically, confidence intervals are discretized, and the proportion of predictions falling within each interval is determined. This method yields a calibration plot in the $[0, 1]$ range, where perfect calibration is represented by a diagonal line.

Calibration performance can be quantified using various metrics, including the miscalibration area (MCA), which is the absolute area under the calibration curve.[50] This metric determines whether a model is systematically overconfident or underconfident. An ideal UQ metric would have an area of zero, while the worst-case scenario would feature an area of 0.5. However, the miscalibration area primarily offers insight into the direction of model mispredictions, not their magnitude. Therefore, a model that is both overconfident and underconfident at different points could potentially attain a perfect score of 0.

Additionally, we report the expected calibration error (ECE) and maximum calibration error (MCE) derived from the calibration curves, which defined as:

$$\text{ECE} = \frac{1}{N} \sum_{i=1}^{N} |\text{CI}_i - \text{EF}_i| \qquad (10)$$

$$\text{MCE} = \max_i (|\text{CI}_i - \text{EF}_i|), \qquad (11)$$

where $N$ is the number of confidence intervals considered, which is set to 100. $\text{CI}_i$ is the $i$-th confidence interval level and $\text{EF}_i$ is the proportion of predictions that fall within that interval.

## 2.6 Uncertainty recalibration

The evaluation of uncertainty calibration on the test dataset demonstrates the initial AutoGNNUQ UQ performance. To refine the UQ performed by AutoGNNUQ, recalibration is conducted using the validation dataset, employing a method similar to that used for computing cNLL. In alignment with the approaches of the benchmark study,[11] which did not implement recalibration, our analysis chooses a comparison using the non-recalibrated version of AutoGNNUQ, as discussed in Section 3.2. However, we still investigate the impact of recalibration on AutoGNNUQ in Section 3.4.2. The recalibration model uses a simple but effective linear scaling approach, aligning with methods in other studies.[51,52] Specifically, we scale the validation uncertainty by a scalar $a$ to minimize the miscalibration area. The same scalar $a$ is applied to test prediction for UQ evaluation, with its optimization conducted in Uncertainty Toolbox.[53]

## 2.7 Computational resources

We used a GPU cluster composed of four nodes, each equipped with an A100 GPU featuring 40 GB of GDDR memory. The software environment used in our experiment was based on Python 3.8 and TensorFlow 2.13.0.[54] For neural architecture search, we used DeepHyper 0.7.0,[37] which employed the TensorFlow-Keras API. For analysis, we used RDKit 2022.3.5, Scikit-learn 1.2.2,[35] and uncertainty-toolbox.[53]

# 3 Results

## 3.1 Neural architecture search

**3.1.1 Search reward.** The efficient discovery of accurate molecular representations is crucial for advancing drug discovery and material design. In this work, we employ the AE search method, which executes 1000 evaluations (small part of architecture space) to discover optimal architectures for predicting molecular properties. The search trajectory of the reward (log-likelihood) with respect to the evaluations is presented in Fig. 5, where a running average with a window size of 50 is applied to minimize noise and produce a smooth visualization of trends in the graphic. The solid line in the graphic represents the average reward accrued over eight different random seeds, with the shaded area denoting the standard deviation. We define search convergence as the point where the reward trajectory reaches a plateau, as observed in Fig. 5.

On average, for all datasets, our analysis revealed that the AE convergence occurs after approximately 400 evaluations. Examples of the optimal and suboptimal architectures discovered during the search are presented in ESI Fig. S1–S8,† which reveal that skip connections and advanced gather layers are more prevalent in high-performing architectures. AutoGNNUQ-simple, trained on atomic numbers and bond orders only, consistently achieves lower search rewards compared to AutoGNNUQ. Post-training loss curves in ESI Fig. S36 and SS7† reveal that AutoGNNUQ-simple records higher average minimum validation losses (NLL) of 1.28, 0.91, 1.16, 0.72, and −0.99 for Lipo, ESOL, FreeSolv, QM7, and QM9, respectively, compared to AutoGNNUQ, which are 1.10, 0.92, 0.85, 0.70, and −1.38. This indicates that the simplest set of features results in reduced model performance. Despite the common practice in NAS of adopting shorter epochs,[37,56] model performance at 30 epochs was assessed against post-training over a full 1000 epochs. As indicated in ESI Fig. S36,† model validation loss for all datasets except QM9 reaches its minimum within 100 epochs. The convergence rate at 30 epochs, defined as the ratio of NLL reduction from epoch 1 to 30 to the reduction from epoch 1 to the minimum NLL, was calculated. For AutoGNNUQ, convergence rates were 0.75, 0.71, 0.65, 0.59, and 0.73 for Lipo, ESOL, FreeSolv, QM7, and QM9, respectively. These findings demonstrate that 30 epochs not only ensure efficient NAS but also provide a reasonably accurate estimate of the potential for each architecture.

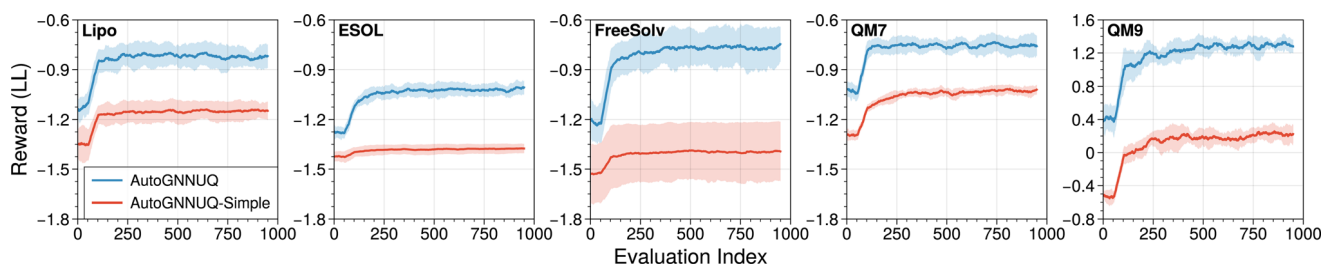

**Fig. 5** Search trajectories for all datasets across eight distinct random seeds. The solid line depicts the average reward while the shaded region indicates the standard deviation. Rewards for all datasets correspond to log–likelihood (LL) values, which we aim to maximize.

**3.1.2 Property prediction performance.** As previously stated, accurate predictions are a critical element of effective UQ. The regression results for Lipo, ESOL, FreeSolv, and QM7 datasets using AutoGNNUQ are shown in Fig. 6, S9–S15,† and Table 1. Additionally, we present results from selecting ten random models at the initialization of AutoGNNUQ to form a random ensemble, as well as results from MC dropout and from the AutoGNNUQ-simple model using the simplest set of features. Prediction parity plots for various random seeds can be found in Fig. S9–S15.† In Fig. 6 and S9–S15,† test set predictions closely align with the diagonal, indicating high accuracy. As illustrated in Table 1, AutoGNNUQ consistently outperforms ensembles of randomly selected models across all datasets, demonstrating that NAS effectively identifies high-performing architectures. AutoGNNUQ surpasses benchmarks in most datasets, with exceptions in ESOL and FreeSolv. This disparity likely stems from their limited sizes (339 and 193 molecules, respectively), which may induce overfitting and reduce prediction accuracy. For all QM9 property predictions, AutoGNNUQ excels, except for the dipole moment ($\mu$). Remarkably, it achieves at least an order of magnitude reduction in the mean absolute error (MAE) for zero-point vibrational energy (ZPVE), internal energy at 0 K ($U_0$), room temperature internal energy ($U$), enthalpy ($H$), and free energy ($G$). MC dropout, using a single AutoGNNUQ architecture with the lowest validation loss, exhibits worse regression performance compared to AutoGNNUQ. This is attributable to the ensemble of diverse architectures in AutoGNNUQ, which mitigates the risk of large errors from individual models.[23] AutoGNNUQ-simple, employing the simplest set of features, achieves the lowest MAE for QM7 and comparable accuracy for Lipo, ESOL, and FreeSolv, while attaining the second-best performance for most QM9 properties. Extended training epochs enable GNNs to derive meaningful molecular representations from even the most basic features.[57] Nonetheless, AutoGNNUQ generally outperforms AutoGNNUQ-simple when using a more comprehensive set of features. It is noteworthy that AutoGNNUQ is not specifically trained to minimize regression error but to optimize NLL. Despite this, its substantial prediction accuracy, particularly when compared with benchmarks, is remarkable and crucial for effective UQ.

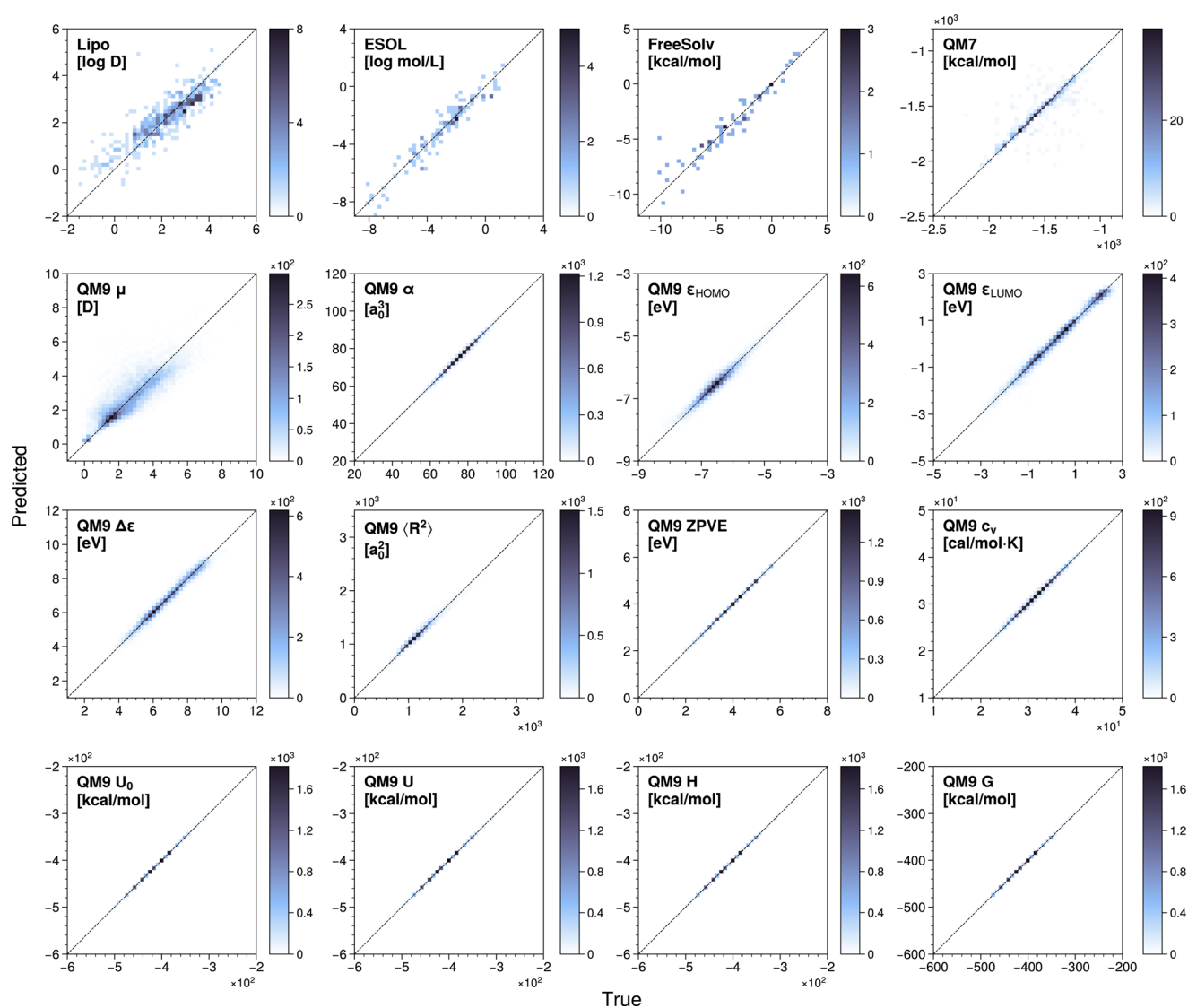

**Fig. 6** Parity plots of all test datasets derived from the same random seed of 0. The diagonal line in each plot represents perfect prediction. The color indicates the number of predictions within each bin.

**3.1.3 Computational time.** The asynchronous NAS approach in AutoGNNUQ is highly parallelizable and scalable, as demonstrated in our previous study using up to sixty compute nodes.[33] In this study, using four compute nodes, we detail the computational costs associated with the entire NAS process and post-training per model in Table 2. By simultaneously performing NAS for improved regression accuracy and UQ, AutoGNNUQ offsets the time required to perform these tasks separately. Notably, even with a large dataset like QM9, exploring 1000 architectures takes less than 35 hours. Increasing the number of compute nodes can further reduce the time required for NAS.

## 3.2 UQ performance

**3.2.1 Negative log-likelihood (NLL).** The results presented in Fig. 7 provide a rigorous evaluation of the ability of the proposed method to accurately estimate variances of normally distributed errors through negative log-likelihood (NLL), which measures the ability of a probabilistic model to predict the likelihood of observed data given a set of parameters.[11] For single-property datasets (Fig. 7a), AutoGNNUQ outperforms the benchmark MPNN ensemble and other UQ methods, with mean NLL values of 0.94, 1.08, 1.64, and 5.62 for Lipo, ESOL, FreeSolv, and QM7, respectively. These values represent reductions of 93%, 92%, 88%, and 70% compared to the benchmark results. We also included QM9 NLL performance, which was not covered in the benchmark study,[11] and compared it with other UQ metrics. For all properties in the QM9 dataset (Fig. 7b), AutoGNNUQ consistently achieves the lowest NLL among all UQ methods. This superior performance is attributed to the architecture search process of AutoGNNUQ, which is explicitly optimized for NLL, allowing effective exploration and identification of high-performing architectures for the UQ task. Among other UQ methods, AutoGNNUQ-simple exhibits the second lowest NLL, followed by MC dropout and the random ensemble. This suggests that optimizing diverse architectures for NLL enhances the final ensemble NLL, whereas a random ensemble of arbitrarily selected architectures fails to achieve comparable performance. Notably, AutoGNNUQ-simple shows substantial variability in NLL, indicating that its minimal feature set leads to unstable UQ performance sensitive to data selection variability. In contrast, AutoGNNUQ demonstrates more consistent NLL values across different random seeds, due to its robust architecture search process, which consistently identifies high-performing architectures.

**3.2.2 Calibrated NLL (cNLL).** Fig. 7 also presents the calibrated negative log-likelihoods (cNLL) for each dataset, where $\hat{\sigma}^2(\mathbf{x}) = a\sigma^2(\mathbf{x}) + b$ is optimized based on the validation data using eqn (8b). The recalibration process aims to identify the best values of $a$ and $b$ that minimize the cNLL. The results indicate that, after recalibration, the benchmark results show a decrease in cNLL values, while the change in AutoGNNUQ cNLL values is not significant, suggesting that AutoGNNUQ is already well calibrated in terms of NLL. Notably, AutoGNNUQ still outperforms the benchmark MPNN ensemble and other UQ methods across most datasets, achieving mean cNLL values of 0.93 for Lipo, 1.07 for ESOL, and 5.61 for QM7. These values indicate reductions of 28%, 38%, and 45% in comparison to the benchmark, respectively. However, for FreeSolv, the mean cNLL is 1.58, which is an 19% increase over the benchmark. This discrepancy could be attributed to the small dataset size, making recalibration on a limited validation set more sensitive and susceptible to overfitting. For all properties in the QM9 dataset, AutoGNNUQ consistently achieves the lowest mean cNLL among all UQ methods. Similarly, AutoGNNUQ-simple has the second lowest average cNLL, followed by MC dropout and the random ensemble. However, AutoGNNUQ-simple still exhibits significant fluctuations in cNLL.

**3.2.3 Miscalibration area.** Fig. 7 also presents the miscalibration area (MCA). It is worth noting that the miscalibration area measures the systematic over- or under-confidence in an aggregated, quantitative sense, rather than the absolute errors of individual compounds. A score of 0 indicates perfect calibration, while a score of 0.5 indicates the worst. A more detailed calibration analysis is provided in Section 3.4.

**Table 2** Computational time for NAS and post-training. Mean value reported with standard deviation in parentheses

| Dataset | NAS time (hours) | Post-training time (minutes) |
|---|---|---|
| Lipo | 6.26 (0.53) | 6.04 (1.12) |
| ESOL | 2.27 (0.06) | 1.62 (0.31) |
| FreeSolv | 2.15 (0.04) | 0.78 (0.14) |
| QM7 | 2.36 (0.26) | 2.26 (0.49) |
| QM9 | 34.69 (1.74) | 149.04 (41.68) |

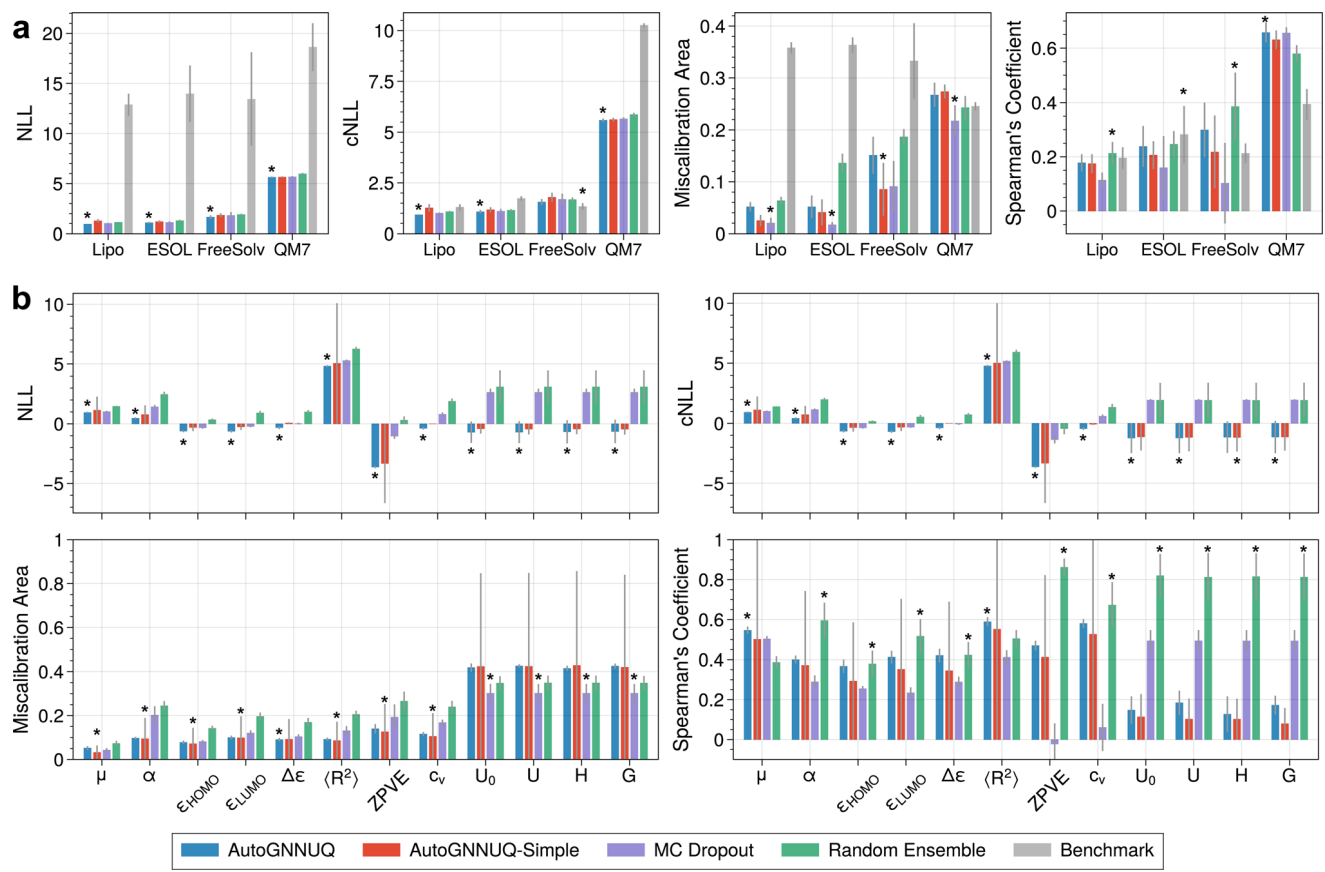

**Fig. 7** Comparison of various methods for UQ performance across (a) Lipo, ESOL, FreeSolv, and QM7 test datasets, and (b) all properties in the QM9 test dataset. The performance metrics evaluated include negative log-likelihood loss (NLL), calibrated NLL (cNLL), miscalibration area, and Spearman's coefficient. The bars represent the mean values of the metrics over eight different runs with different random seeds for test data splitting, while the error bars indicate the standard deviations. An asterisk denotes the best-performing model.

AutoGNNUQ surpasses the benchmark MPNN ensemble on most datasets, shown by mean MCA values of 0.052, 0.052, and 0.15 for Lipo, ESOL, and FreeSolv, respectively. This equates to an 86%, 86%, and 55% reduction in comparison to the benchmark results. However, for QM7, MCA of AutoGNNUQ is 0.27, 9% higher than the benchmark. For all datasets, including QM9, AutoGNNUQ-simple and MC dropout exhibit the lowest mean MCA followed by AutoGNNUQ and the random ensemble. The higher MCA of AutoGNNUQ is attributed to the overestimation of aleatoric uncertainty, as discussed in Section 3.4, where recalibrated MCA values are also presented.

**3.2.4 Spearman's rank correlation coefficient.** Fig. 7 displays the Spearman's rank correlation coefficient (Spearman's coefficient), which indicates the relationship between uncertainty and prediction error. A score of 1 indicates a positive correlation between uncertainty and prediction error, while a score of −1 indicates a negative correlation, and a score of 0 indicates no correlation. For the Lipo and ESOL datasets, AutoGNNUQ achieves mean coefficients of 0.18 and 0.24, respectively, marking decreases of 9% and 16% compared to the benchmark results. In contrast, for FreeSolv and QM7, AutoGNNUQ has mean coefficients of 0.30 and 0.66, indicating increases of 41% and 67% over the benchmark values, respectively. For all datasets, including QM9, the random ensemble exhibits the highest mean coefficients, followed by AutoGNNUQ, AutoGNNUQ-simple, and MC dropout.

However, we should note that higher uncertainty values do not necessarily indicate high prediction error. A well-calibrated model that captures underlying noise in data can perform well despite high uncertainty. Moreover, limited data in certain input spaces can make it challenging to estimate the true level of uncertainty, leading to poor model performance despite low uncertainty and high error. A closer examination of the relationship between error and uncertainty is presented in Section 3.3.

**3.2.5 Confidence curve.** In addition to Spearman's coefficient, another ranking-based UQ metric is the confidence curve, which is illustrated in ESI Fig. S32–S35,† showing the confidence curves for all evaluated UQ methods. For all datasets, the total uncertainty exhibits an overall decreasing trend, which is expected since removing data points with the highest uncertainty should result in data with lower associated errors. However, smaller datasets such as Lipo, ESOL, and FreeSolv have noisier curves and less consistently decreasing slopes. Table S7† provides a comparison of the AUCO values across different methods, which describe area between the oracle confidence curve and the observed confidence curve. Except for the FreeSolv dataset, where the random ensemble achieves the lowest AUCO, AutoGNNUQ achieves the lowest AUCO across all other datasets, including all properties in QM9, indicating superior performance.

### 3.3 Prediction error and uncertainty

Fig. 8 and S16–S22† illustrate the relationship between prediction error and uncertainty, measured as standard deviation (std). For Lipo, ESOL, FreeSolv, and QM7, the majority of observed errors fall within one std, with percentages of $75.9 \pm 1.2\%$, $75.8 \pm 3.2\%$, $85.5 \pm 4.6\%$, and $90.9 \pm 1.0\%$, respectively, across eight random seeds. A small fraction of predictions exceed two std thresholds, specifically $3.2 \pm 0.5\%$, $3.3 \pm 1.0\%$, $3.4 \pm 1.4\%$, and $3.6 \pm 0.8\%$ for Lipo, ESOL, FreeSolv, and QM7, respectively. For all QM9 properties, as shown in ESI Fig. S16–S22,† the majority of observed errors fall within one std, with the lowest percentage being $75.8\% \pm 1.1\%$. Errors exceeding two std are all below $2.7\% \pm 0.3\%$.

Uncertainty calibration is achieved when the uncertainty estimates accurately align with the actual variability of the predictions. The uncertainty estimates from the model indicate reasonable calibration for Lipo and ESOL. Under the assumption of a Gaussian distribution for prediction errors, approximately 68% of these errors are expected to fall within one std. from the mean prediction, and 95% within two std.[58] Remarkably, the one std. ratio is close to 90% for FreeSolv and QM7, and nearly 100% for $U_0$, $U$, $H$, and $G$ in QM9, which indicates that the uncertainty estimates for these datasets are notably conservative.

### 3.4 Confidence-based calibration

To further examine whether a model is well-calibrated, we show the confidence-based calibration curves as in Fig. 9a. We observe that for total uncertainty in Lipo and ESOL, the confidence curves align closely with the diagonal, indicating good calibration but a slight underconfidence, which means the uncertainty in predictions is overestimated. For FreeSolv, QM7, and the QM9 properties $U_0$, $U$, $H$, and $G$, the predictions exhibit even greater underconfidence. Clearly, aleatoric uncertainty significantly contributes to the overestimated uncertainty. To identify the source of this underconfidence, we look into the decomposition of uncertainty in the following section.

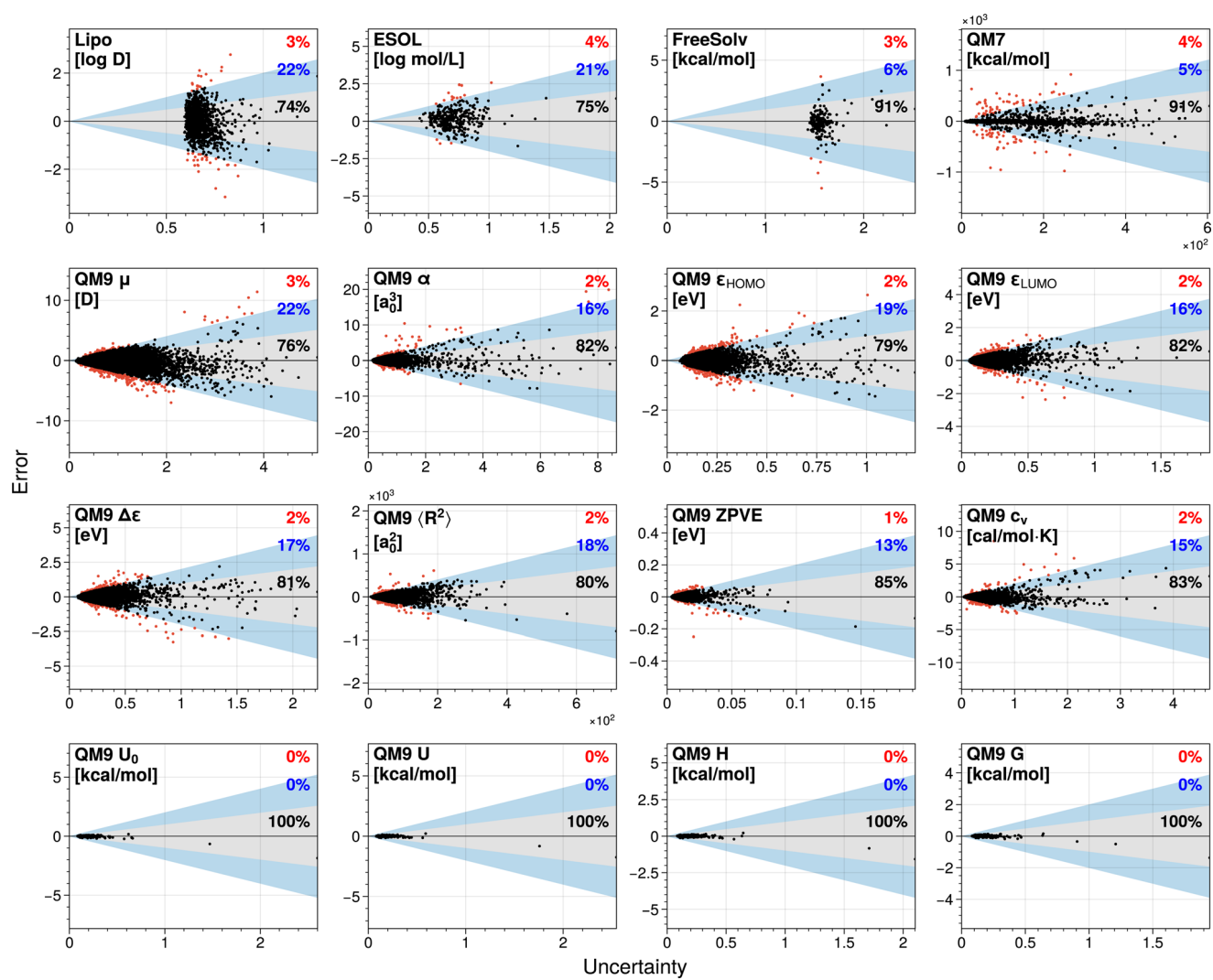

**Fig. 8** Relationship between model error and uncertainty of all test datasets derived from the same random seed of 0. The gray region denotes to one standard deviation (std) and blue to two std. The points with model errors that fall within either of these two bounds are shown in black, and the percentage within the gray or blue regions is annotated in black and blue, respectively, in each graph. The points exceeding two std are marked in red, with the proportion of such points annotated on each graph in red.

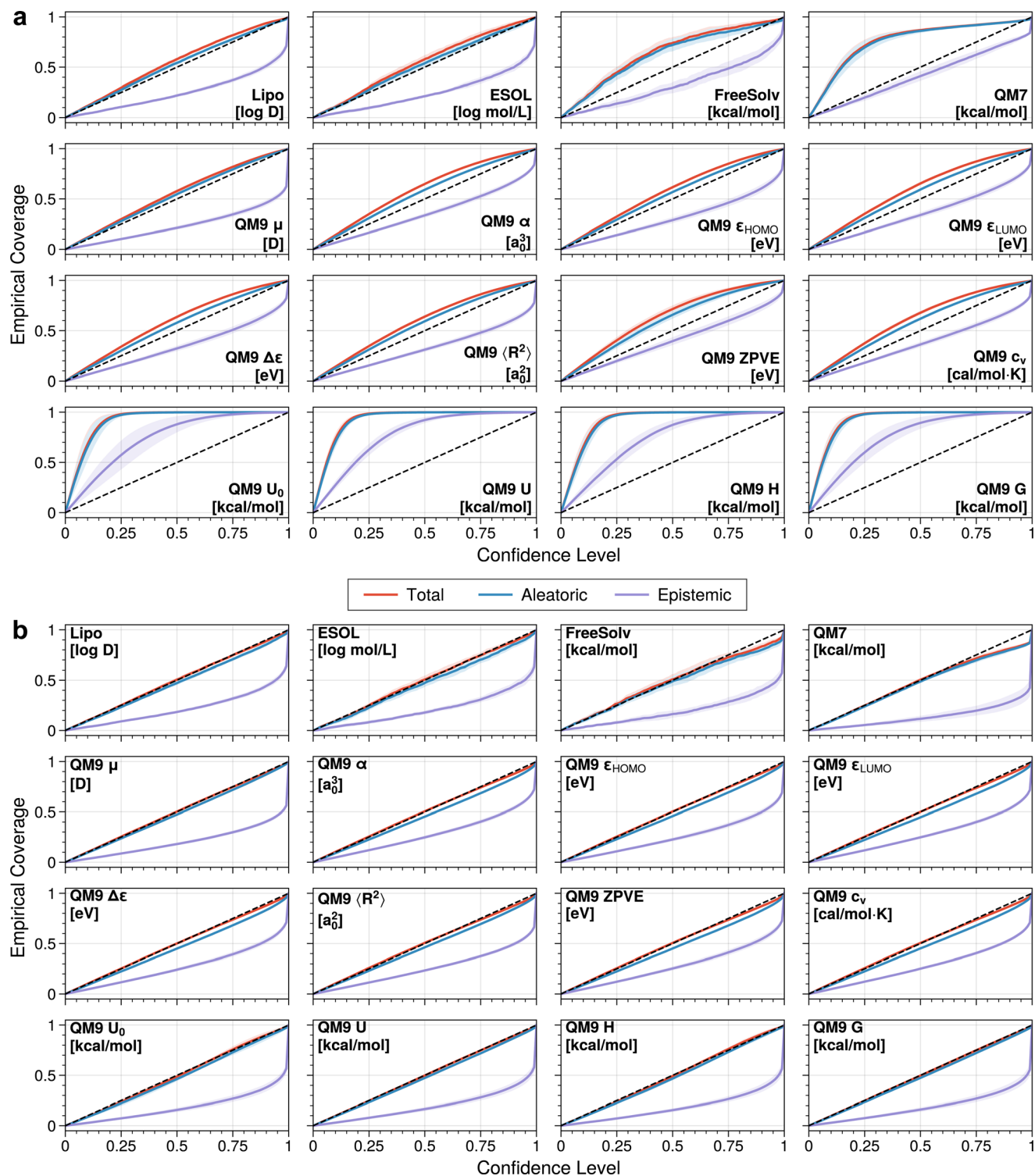

**Fig. 9** Confidence-based calibration curves for all datasets across eight distinct random seeds (a) before and (b) after recalibration. The confidence level represents the predicted probability of obtaining the correct error value within a given percentile. Empirical coverage reflects the actual probabilities. The diagonal line shows the behavior of a perfectly calibrated model. Solid lines represent mean calibration curves for total, epistemic, and aleatoric uncertainties across eight random seeds, while shaded areas indicate standard deviation.

**3.4.1 Uncertainty decomposition.** The decomposition of uncertainty into aleatoric and epistemic uncertainties is a crucial aspect of statistical modeling and decision-making. As presented in eqn (5b), aleatoric uncertainty is determined by the mean value of the predictive variance of each model in the ensemble, whereas epistemic uncertainty is determined by the predictive variance of the mean value of each model in the ensemble. Further visualized in Fig. 10, the cumulative density distribution of aleatoric and epistemic uncertainties reveals that, in all datasets, epistemic uncertainty is smaller than aleatoric uncertainty, which is indicated by the dominance of the blue curve over the red curve.

However, recent studies indicate that ensemble-based models tend to overestimate aleatoric uncertainty.[52] The ensemble model, deriving its mean by averaging outcomes from multiple models, should theoretically exhibit reduced error, leading to lower aleatoric uncertainty than individual models.[59] Averaging predictions of aleatoric uncertainty from individual models, as described in eqn (5b), typically does not decrease the magnitude of aleatoric uncertainty. This approach leads to overestimation of aleatoric uncertainty and results in an underconfident ensemble model,[59] which is shown in Fig. 9a.

**3.4.2 Uncertainty recalibration.** After identifying aleatoric uncertainty as the primary source of overestimation uncertainty, we propose recalibration based on validation predictions. Specifically, we multiply aleatoric uncertainty, $\sigma_{alea}$, by a scalar $a$ to minimize validation MCA. The scaling factor $a$ adjusts test aleatoric and epistemic uncertainties jointly to avoid biased total uncertainties.[51] We should note that uncertainty recalibration can be customized to optimize various objectives, including NLL, MCA, and ECE. The decision to recalibrate with respect to MCA aims to mitigate the significant overestimation of aleatoric uncertainty.

The recalibration results are shown in Fig. 9b, which illustrates that after recalibration, both total and aleatoric uncertainty curves more closely follow the diagonal line, indicating refined calibration. The level of epistemic uncertainty remains significantly lower, resulting in a minimal impact on the total uncertainty. The scaling ratios, listed in ESI Table S2,† are all below 1, indicating a substantial decrease in uncertainties across all datasets. Notably, QM7 records an average ratio below 0.3, while the QM9 properties $U_0$, $U$, $G$, and $H$ exhibit ratios below 0.15, showing the most significant reduction in scaling.

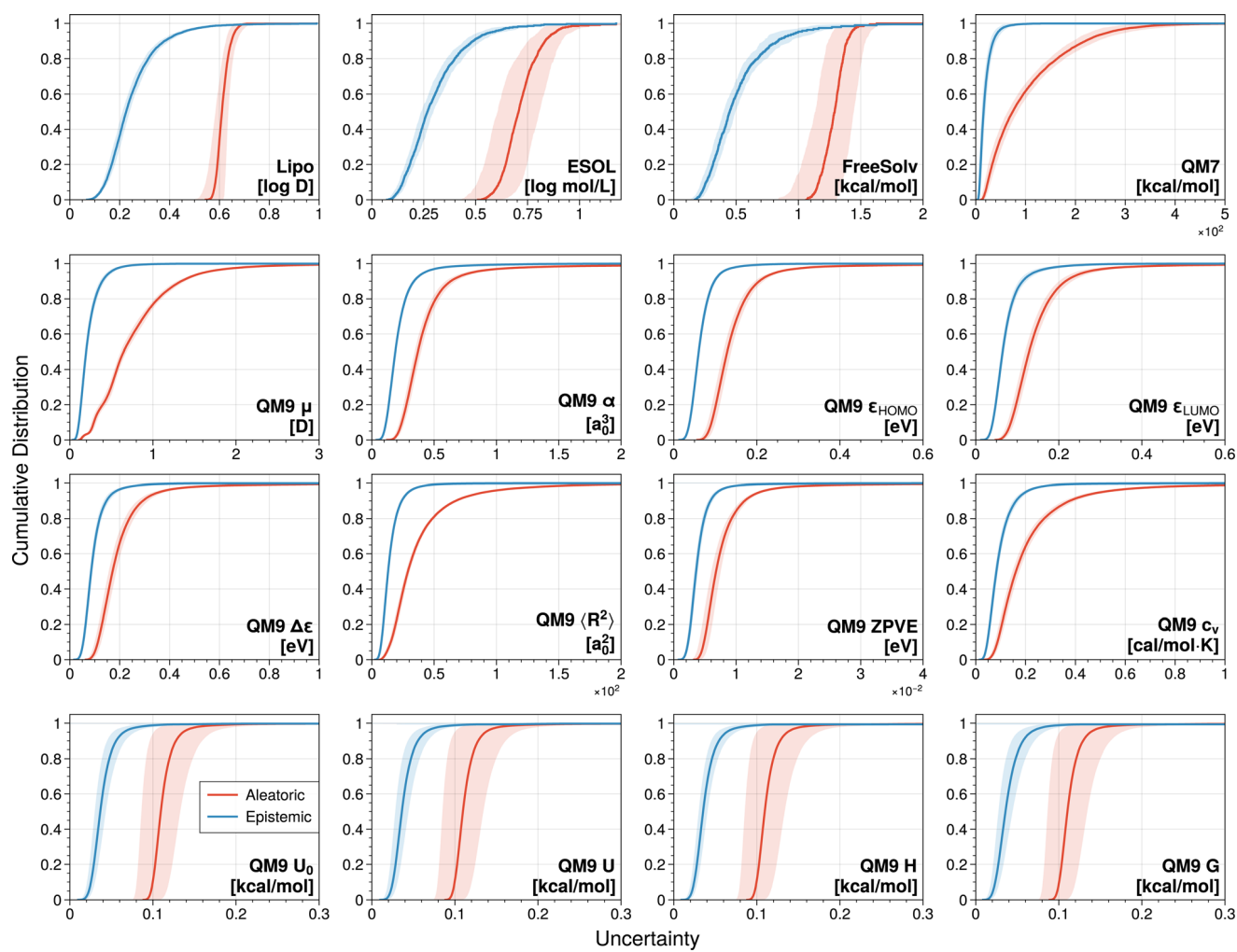

**Fig. 10** Cumulative density distribution of uncertainty decomposition of all test datasets across eight random seeds. The solid line within the distribution represents the mean, while the shaded area signifies the standard deviation. For all datasets, the curve representing epistemic uncertainty is above the curve of aleatoric uncertainty, indicating that most of the uncertainties in the predictions arise from aleatoric uncertainty.

**Table 3** Confidence-based calibration metrics of AutoGNNUQ before and after recalibration based on the validation dataset. Mean value reported with standard deviation in parentheses. The best result is bold

| Dataset | Property | ECE | | MCE | | MCA | |
|---|---|---|---|---|---|---|---|
| | | Before | After | Before | After | Before | After |
| Lipo | $\log P$ | 0.051 (0.009) | **0.011 (0.005)** | 0.089 (0.014) | **0.028 (0.008)** | 0.052 (0.009) | **0.011 (0.005)** |
| ESOL | $S_{water}$ | 0.051 (0.022) | **0.024 (0.008)** | 0.099 (0.031) | **0.054 (0.014)** | 0.051 (0.022) | **0.024 (0.008)** |
| FreeSolv | $\Delta G_{hyd}$ | 0.150 (0.035) | **0.036 (0.009)** | 0.266 (0.056) | **0.089 (0.018)** | 0.151 (0.036) | **0.036 (0.009)** |
| QM7 | $\Delta H_{atom}$ | 0.265 (0.023) | **0.028 (0.006)** | 0.477 (0.042) | **0.118 (0.013)** | 0.268 (0.023) | **0.029 (0.006)** |
| QM9 | $\mu$ | 0.052 (0.009) | **0.005 (0.002)** | 0.082 (0.014) | **0.012 (0.005)** | 0.052 (0.009) | **0.005 (0.002)** |
| | $\alpha$ | 0.096 (0.005) | **0.008 (0.002)** | 0.152 (0.009) | **0.024 (0.004)** | 0.097 (0.005) | **0.008 (0.003)** |
| | $\varepsilon_{HOMO}$ | 0.077 (0.008) | **0.006 (0.001)** | 0.120 (0.011) | **0.016 (0.003)** | 0.078 (0.008) | **0.006 (0.001)** |
| | $\varepsilon_{LUMO}$ | 0.099 (0.009) | **0.008 (0.002)** | 0.156 (0.014) | **0.025 (0.006)** | 0.100 (0.009) | **0.008 (0.002)** |
| | $\Delta\varepsilon$ | 0.089 (0.008) | **0.008 (0.002)** | 0.140 (0.012) | **0.023 (0.005)** | 0.090 (0.008) | **0.008 (0.002)** |
| | $\langle R^2 \rangle$ | 0.091 (0.008) | **0.010 (0.002)** | 0.141 (0.013) | **0.025 (0.005)** | 0.092 (0.008) | **0.010 (0.002)** |
| | ZPVE | 0.138 (0.021) | **0.010 (0.002)** | 0.217 (0.033) | **0.030 (0.004)** | 0.140 (0.021) | **0.010 (0.002)** |
| | $c_v$ | 0.115 (0.009) | **0.010 (0.002)** | 0.180 (0.016) | **0.030 (0.006)** | 0.116 (0.010) | **0.011 (0.002)** |
| | $U_0$ | 0.414 (0.019) | **0.014 (0.015)** | 0.757 (0.030) | **0.029 (0.023)** | 0.419 (0.019) | **0.014 (0.015)** |
| | $U$ | 0.422 (0.008) | **0.006 (0.002)** | 0.768 (0.017) | **0.016 (0.007)** | 0.426 (0.008) | **0.006 (0.002)** |
| | $H$ | 0.411 (0.012) | **0.012 (0.005)** | 0.749 (0.021) | **0.026 (0.008)** | 0.415 (0.012) | **0.012 (0.006)** |
| | $G$ | 0.421 (0.011) | **0.007 (0.002)** | 0.767 (0.022) | **0.019 (0.005)** | 0.426 (0.011) | **0.007 (0.003)** |

UQ calibration metrics pre- and post-recalibration for AutoGNNUQ are detailed in Table 3. Notably, the ECE, MCE, and MCA for total uncertainty significantly decrease following recalibration. This indicates that UQ calibration performance of AutoGNNUQ can be further refined by recalibration. Similar decreases are observed for AutoGNNUQ-simple, MC dropout, and the random ensemble (ESI Table S2†). For all UQ methods, post-recalibration results for Lipo, ESOL, FreeSolv, and QM7 are comparable. However, for all properties in QM9, the random ensemble performs worse, while other methods yield similar results (ESI Tables S3–S5†).

### 3.5 Molecular uncertainties

To address the presence of reducible epistemic uncertainties in AutoGNNUQ, we investigate the molecular characteristics and functional groups that contribute to these uncertainties. By analyzing these factors, we have a potential to improve the representation of the molecules and subsequently reduce the uncertainties. To accomplish this, we use MACCS (Molecular ACCess System) keys,[44] which are binary fingerprints that represent each molecule in our dataset by encoding the presence or absence of particular functional groups and substructures. These functional groups include characteristics, such as actinide (index 4), hydroxyl group (index 139), multiple nitrogen atoms (index 142). By analyzing the contribution of each functional group to high uncertainties, we can gain a better understanding of the impact of individual groups on molecular properties and improve the overall representation of the molecules.

We use t-SNE (t-Distributed Stochastic Neighbor Embedding)[60] to visualize the relationship between the MACCS key

vectors and the molecules. The t-SNE algorithm creates probability distributions for the similarities between the high-dimensional data points and the low-dimensional space, optimizing the latter to closely match the former. By mapping the 166-dimensional vectors of all molecules into two dimensions using t-SNE, we can create a plot that can reveal patterns and clusters within complex datasets. This enables us to identify relationships between the molecules and the functional groups that make up their MACCS key vectors. In Fig. 11a, we show the t-SNE plot of the QM7 dataset (random seed 0) and K-means clustering. The optimal number of clusters K in K-means clustering is determined using silhouette scores† Which evaluate how similar a data point is to its own cluster compared to other clusters. Higher average silhouette scores indicate better-defined clusters, and the K associated with the highest silhouette score is selected. In Fig. 11b, we show the t-SNE plot of the QM7 dataset with associated epistemic uncertainties. It is important to note that the empty spaces in t-SNE plots do not indicate areas of sparse data. These gaps result from the algorithm optimization process, which aims to preserve local data structures in lower-dimensional representations, leading to non-uniform point distributions. The results presented in Fig. 11 indicate that molecules with low epistemic uncertainties are predominantly comprised of O-heterocycles, ether, and hydroxyl groups, while those with high epistemic uncertainties are largely composed of N-heterocycles and sulfonyl groups. This suggest that it is worthwhile to take a closer look at molecules containing N-heterocycles and sulfonyl groups to determine if there are any limitations with data representations. Additional t-SNE results for various datasets are provided in ESI Fig. S23 through S31.†

### 3.6 Out-of-distribution performance

Accurate OOD property prediction and robust UQ are crucial for optimally selecting new molecules for characterization. The PC9 and QM9 molecular datasets exhibit significant differences.

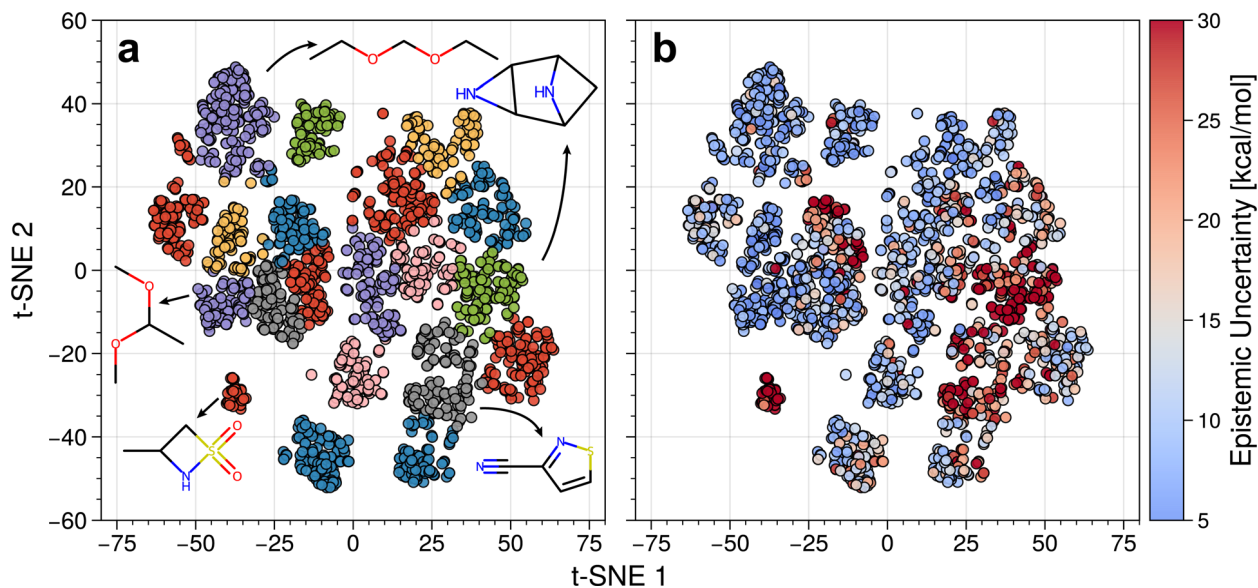

**Fig. 11** t-SNE plot of the MACCS keys of the test dataset. (a) Clustering by $k$-means algorithm with $k = 18$. (b) Epistemic uncertainties (standard deviation) associated with each molecule. Low uncertainties are represented in blue and high uncertainties in red.

PC9 uniquely contains molecules with multiplicities greater than one, including radicals and triplets. Additionally, the first principal component (PC) of MACCS fingerprints (Fig. 12a) reveals different distributions, indicating variations in functional groups and chemical substructures. OOD property prediction for $\varepsilon_{\text{HOMO}}$ and $\varepsilon_{\text{LUMO}}$ (Fig. 12d) shows high accuracy, with MAEs of 0.340 and 0.369 eV, respectively. For molecules with a multiplicity ($m_s$) of 1, where all electrons are paired, the MAEs for $\varepsilon_{\text{HOMO}}$ and $\varepsilon_{\text{LUMO}}$ are 0.299 and 0.358 eV. For radicals with $m_s = 2$, the MAEs increase to 0.888 and 0.493 eV. For triplets with $m_s = 3$, the MAEs are 0.916 and 0.619 eV. Since all QM9 molecules have $m_s = 1$, PC9 molecules with $m_s = 1$ are expected to exhibit higher prediction accuracy compared to those with higher multiplicities, which are not represented in the QM9 dataset. Furthermore, the cumulative distributions of aleatoric and epistemic uncertainty (Fig. 12b) in PC9 show a predominance of epistemic uncertainty, contrasting with prior in-distribution studies where aleatoric uncertainty is more prevalent. This heightened epistemic uncertainty is attributed to the inadequate representation of distributions in the training data for OOD molecules. The confidence-based calibration (Fig. 12c) illustrates that the UQ provided by AutoGNNUQ is well-calibrated for OOD molecules, with MCA of 0.010 for $\varepsilon_{HOMO}$ and 0.061 for $\varepsilon_{LUMO}$. Additionally, as depicted in Fig. S38,† 85% and 87% of prediction errors fall within two std. for $\varepsilon_{HOMO}$ and $\varepsilon_{LUMO}$, respectively. This proportion is lower than the expected 95% for a Gaussian distribution, indicating a slight underestimation of uncertainty.

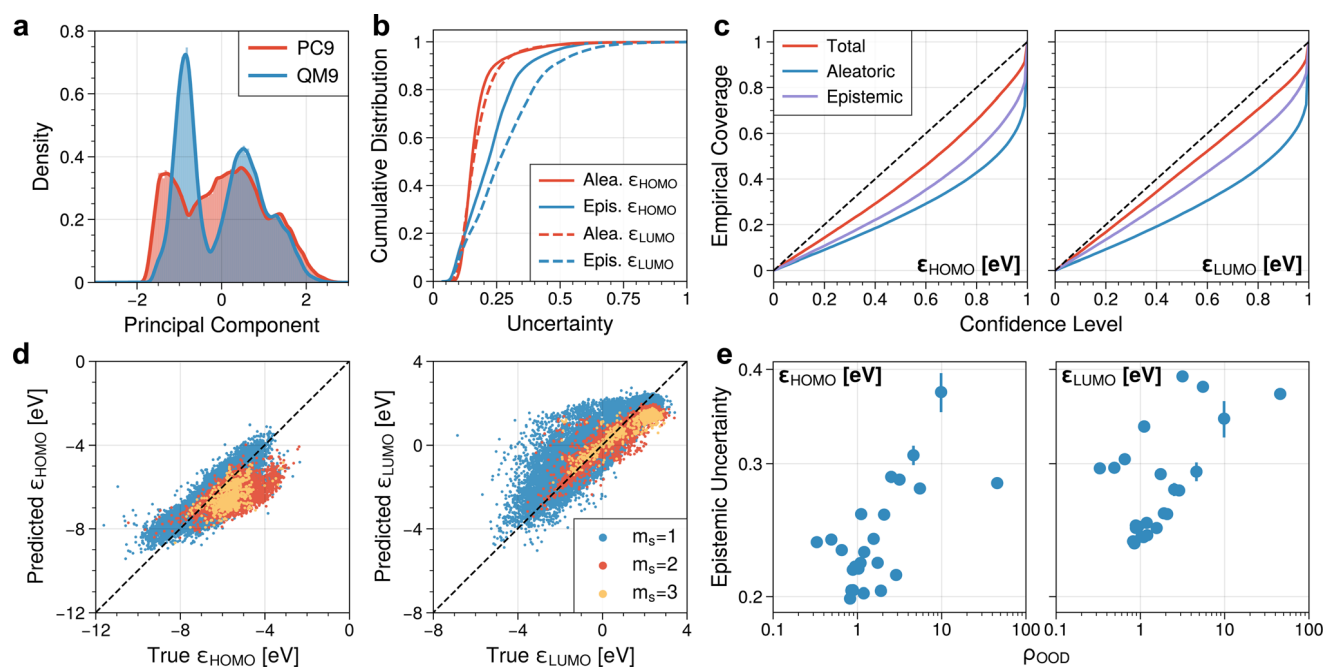

**Fig. 12** Performance of AutoGNNUQ trained on QM9 for prediction and UQ on OOD PC9 molecules. (a) PCA of PC9 and QM9 molecules based on MACCS fingerprints, showing the first PC. (b) Cumulative density distribution of uncertainty decomposition for PC9 molecules. (c) Confidence-based calibration curves for PC9 predictions. (d) Parity plots of property prediction for PC9 molecules with varying multiplicities ($m_s$). (e) Relationship between epistemic uncertainty and degree of OOD status ($\rho_{\text{OOD}}$) for each bin from the first PC of MACCS fingerprints (log scale). Each dot represents the mean epistemic uncertainty, with error bars indicating the standard error of the mean.

We further investigate the relationship between the degree of OOD status and epistemic uncertainty. We define a descriptor, $\rho_{OOD}$, to quantify the OOD status of groups of molecules using the first PC of MACCS fingerprints. The distribution in Fig. 12a is divided into 25 evenly spaced bins. Each bin includes molecules with similar functional groups and chemical substructures. For bin $i$, the OOD status $\rho_{OOD,i}$ is defined as the ratio of the number of molecules in the $i$-th bin for PC9 to those in QM9, i.e., $\rho_{OOD,i} = \rho_{PC9,i}/\rho_{QM9,i}$. A high $\rho_{OOD,i}$ indicates that PC9 molecules are underrepresented in the QM9 dataset. Fig. 12e shows a positive correlation between average epistemic uncertainty and $\rho_{OOD}$ for each bin, with Spearman's coefficients of 0.561 and 0.462 for $\varepsilon_{LUMO}$ and $\varepsilon_{HOMO}$, respectively. This suggests that molecules with a high degree of OOD status are effectively characterized by their epistemic uncertainties. In addition, for PC7 molecules with $m_s = 1$, the epistemic uncertainties are $0.227 \pm 0.117$ and $0.279 \pm 0.153$ for $\varepsilon_{LUMO}$ and $\varepsilon_{HOMO}$, respectively. These values are lower than those for molecules with $m_s > 1$, which exhibit epistemic uncertainties of $0.229 \pm 0.082$ and $0.308 \pm 0.102$. This further demonstrates that epistemic uncertainty effectively captures the underrepresentation of higher multiplicity molecules. Overall, the high prediction accuracy and robust UQ performance on OOD molecules underscore the potential of AutoGNNUQ for applications in uncertainty-guided active learning and experimental design.

## 4 Conclusions

In summary, this paper presents AutoGNNUQ, a novel technique for uncertainty quantification in machine learning models used for molecular prediction. AutoGNNUQ leverages an aging evolution approach to construct a diverse graph neural network ensemble that models epistemic uncertainty while preserving aleatoric uncertainty quality. Our experiments on several benchmark datasets demonstrate that AutoGNNUQ outperforms existing methods in terms of prediction accuracy and uncertainty quantification. By decomposing aleatoric and epistemic uncertainty and performing recalibration, we gain valuable insights into areas for reducing uncertainty and improving uncertainty calibration, and t-SNE visualization enhances our understanding of the correlation between molecular features and uncertainties. The accurate property prediction and robust uncertainty quantification on out-of-distribution data further highlight the potential of AutoGN-NUQ for future applications in active learning for materials discovery.

Moving forward, we plan to delve into the specific causes of epistemic uncertainty, including imbalanced data, insufficient representations, and poor models. We intend to expand our approach to tackle uncertainty quantification in classification problems and other domains beyond molecular representations. Despite optimizing only negative log-likelihood, we achieve high regression accuracy, especially for QM9. In the future, we plan to optimize both prediction accuracy and uncertainty quantification as a multi-objective search to better understand their interaction. Additionally, we aim to incorporate the evidential model with neural architecture search to perform uncertainty quantification and decomposition from a different perspective. Overall, AutoGNNUQ presents a promising direction for improving the accuracy and reliability of machine learning models in various applications. The results of this study highlight the potential of AutoGNNUQ to advance the field of uncertainty quantification in machine learning models, particularly in molecular prediction.

## Data availability

The data and code used to generate results are publicly available at **https://github.com/sjiang87/deephyper**.

## Author contributions

S. J. developed the concept of AutoGNNUQ, implemented it, set up and carried out the computational experiments including formal analysis and visualization, and wrote the original draft of the manuscript. S. Q. supported the development of the computational experiments and the analysis of the results, and edited the manuscript. R. C. V. L., P. B., and V. M. Z. acquired the funding, supervised the work, and edited the manuscript.

## Conflicts of interest

There are no conflicts to declare.

## Acknowledgements

This material is based on work supported by the U.S. Department of Energy (DOE), Office of Science, Office of Advanced Scientific Computing Research, under Contract DE-AC02-06CH11357. We also acknowledge partial funding from the U.S. National Science Foundation (NSF) under BIGDATA grant IIS-1837812.

## References

1 A. Cherkasov, E. N. Muratov, D. Fourches, A. Varnek, I. I. Baskin, M. Cronin, J. Dearden, P. Gramatica,

Y. C. Martin, R. Todeschini, *et al.*, *J. Med. Chem.*, 2014, **57**, 4977–5010.

2 W. P. Walters and R. Barzilay, *Acc. Chem. Res.*, 2020, **54**, 263–270.

3 E. N. Feinberg, D. Sur, Z. Wu, B. E. Husic, H. Mai, Y. Li, S. Sun, J. Yang, B. Ramsundar and V. S. Pande, *ACS Cent. Sci.*, 2018, **4**, 1520–1530.

4 Z. Hao, C. Lu, Z. Huang, H. Wang, Z. Hu, Q. Liu, E. Chen and C. Lee, *Proceedings of the 26th ACM SIGKDD International Conference on Knowledge Discovery & Data Mining*, 2020, pp. 731–752.

5 J. Gilmer, S. S. Schoenholz, P. F. Riley, O. Vinyals and G. E. Dahl, *International conference on machine learning*, 2017, pp. 1263–1272.

6 S. Qin, S. Jiang, J. Li, P. Balaprakash, R. C. Van Lehn and V. M. Zavala, *Digital Discovery*, 2023, **2**, 138–151.

7 Y. Gal and Z. Ghahramani, *International conference on machine learning*, 2016, pp. 1050–1059.

8 A. F. Psaros, X. Meng, Z. Zou, L. Guo and G. E. Karniadakis, *J. Comput. Phys.*, 2023, 111902.

9 A. Lysenko, A. Sharma, K. A. Boroevich and T. Tsunoda, *Life Sci. Alliance*, 2018, **1**, 1–14.

10 J. Gawlikowski, C. R. N. Tassi, M. Ali, J. Lee, M. Humt, J. Feng, A. Kruspe, R. Triebel, P. Jung, R. Roscher, *et al.*, *arXiv*, 2021, preprint, arXiv:2107.03342, DOI: **10.48550/arXiv.2107.03342**.

11 L. Hirschfeld, K. Swanson, K. Yang, R. Barzilay and C. W. Coley, *J. Chem. Inf. Model.*, 2020, **60**, 3770–3780.

12 D. A. Nix and A. S. Weigend, *Proceedings of 1994 ieee international conference on neural networks (ICNN'94)*, 1994, pp. 55–60.

13 Y. Gal and Z. Ghahramani, *arXiv*, 2015, preprint, arXiv:1506.02158, DOI: **10.48550/arXiv.1506.02158**.

14 Y. Gal, J. Hron and A. Kendall, *Neural Information Processing Systems*, 2017.

15 S. Jain and S. P. K., *Proceedings of the 6th Joint International Conference on Data Science & Management of Data (10th ACM IKDD CODS and 28th COMAD)*, 2023, pp. 138.

16 R. Grosse and J. Martens, *International Conference on Machine Learning*, 2016, pp. 573–582.

17 H. Ritter, A. Botev and D. Barber, *6th International Conference on Learning Representations, ICLR 2018-Conference Track Proceedings*, 2018.

18 J. Lee, M. Humt, J. Feng and R. Triebel, *International Conference on Machine Learning*, 2020, pp. 5702–5713.

19 L. K. Hansen and P. Salamon, *IEEE Trans. Pattern Anal. Mach. Intell.*, 1990, **12**, 993–1001.

20 B. Lakshminarayanan, A. Pritzel and C. Blundell, Simple and scalable predictive uncertainty estimation using deep ensembles, *Neural Information Processing Systems*, 2017, pp. 6405–6416.

21 I. E. Livieris, L. Iliadis and P. Pintelas, *Evol. Syst.*, 2021, **12**, 155–167.

22 E. J. Herron, S. R. Young and T. E. Potok, *International Conference on High Performance Computing*, 2020, pp. 223–234.

23 R. Egele, R. Maulik, K. Raghavan, P. Balaprakash and B. Lusch, *arXiv*, 2021, preprint, arXiv:2110.13511, DOI: **10.48550/arXiv.2110.13511**.

24 E. Real, A. Aggarwal, Y. Huang and Q. V. Le, *Proceedings of the AAAI Conference on Artificial Intelligence*, 2019, pp. 4780–4789.

25 Z. Wu, B. Ramsundar, E. N. Feinberg, J. Gomes, C. Geniesse, A. S. Pappu, K. Leswing and V. Pande, *Chem. Sci.*, 2018, **9**, 513–530.

26 D. Mendez, A. Gaulton, A. P. Bento, J. Chambers, M. De Veij, E. Félix, M. P. Magariños, J. F. Mosquera, P. Mutowo, M. Nowotka, *et al.*, *Nucleic Acids Res.*, 2019, **47**, D930–D940.

27 D. L. Mobley and J. P. Guthrie, *J. Comput.-Aided Mol. Des.*, 2014, **28**, 711–720.

28 J. S. Delaney, *J. Chem. Inf. Comput. Sci.*, 2004, **44**, 1000–1005.

29 L. C. Blum and J.-L. Reymond, *J. Am. Chem. Soc.*, 2009, **131**, 8732–8733.

30 G. Montavon, M. Rupp, V. Gobre, A. Vazquez-Mayagoitia, K. Hansen, A. Tkatchenko, K.-R. Müller and O. A. Von Lilienfeld, *New J. Phys.*, 2013, **15**, 095003.

31 R. Ramakrishnan, P. O. Dral, M. Rupp and O. A. Von Lilienfeld, *Sci. Data*, 2014, **1**, 1–7.

32 L. Ruddigkeit, R. Van Deursen, L. C. Blum and J.-L. Reymond, *J. Chem. Inf. Model.*, 2012, **52**, 2864–2875.

33 S. Jiang and P. Balaprakash, *2020 IEEE International conference on big data (big data)*, 2020, pp. 1346–1353.

34 P. Veličković, G. Cucurull, A. Casanova, A. Romero, P. Lio and Y. Bengio, *arXiv*, 2017, preprint, arXiv:1710.10903, DOI: **10.48550/arXiv.1710.10903**.

35 K. Xu, W. Hu, J. Leskovec and S. Jegelka, *arXiv*, 2018, preprint, arXiv:1810.00826, DOI: **10.48550/arXiv.1810.00826**.

36 D. Grattarola and C. Alippi, *IEEE Comput. Intell. Mag.*, 2021, **16**, 99–106.

37 P. Balaprakash, M. Salim, T. D. Uram, V. Vishwanath and S. M. Wild, *2018 IEEE 25th International Conference on High Performance Computing (HiPC)*, 2018, pp. 42–51.

38 R. Maulik, R. Egele, B. Lusch and P. Balaprakash, *SC20: International Conference for High Performance Computing, Networking, Storage and Analysis*, 2020, pp. 1–14.

39 D. P. Kingma and J. Ba, *arXiv*, 2014, preprint, arXiv:1412.6980, DOI: **10.48550/arXiv.1412.6980**.

40 A. Amini, W. Schwarting, A. Soleimany and D. Rus, *Adv. Neural Inf. Process. Syst.*, 2020, **33**, 14927–14937.

41 A. P. Soleimany, A. Amini, S. Goldman, D. Rus, S. N. Bhatia and C. W. Coley, *ACS Cent. Sci.*, 2021, **7**, 1356–1367.

42 T. Yin, G. Panapitiya, E. D. Coda and E. G. Saldanha, *J. Cheminf.*, 2023, **15**, 105.

43 M. Glavatskikh, J. Leguy, G. Hunault, T. Cauchy and B. Da Mota, *J. Cheminf.*, 2019, **11**, 1–15.

44 J. L. Durant, B. A. Leland, D. R. Henry and J. G. Nourse, *J. Chem. Inf. Comput. Sci.*, 2002, **42**, 1273–1280.

45 J. P. Janet, C. Duan, T. Yang, A. Nandy and H. J. Kulik, *Chem. Sci.*, 2019, **10**, 7913–7922.

46 E. Ilg, O. Cicek, S. Galesso, A. Klein, O. Makansi, F. Hutter and T. Brox, *Proceedings of the European Conference on Computer Vision (ECCV)*, 2018, pp. 652–667.

47 G. Scalia, C. A. Grambow, B. Pernici, Y.-P. Li and W. H. Green, *J. Chem. Inf. Model.*, 2020, **60**, 2697–2717.

48 F. K. Gustafsson, M. Danelljan and T. B. Schon, *Proceedings of the IEEE/CVF Conference on Computer Vision and Pattern Recognition Workshops*, 2020, pp. 318–319.

49 V. Kuleshov, N. Fenner and S. Ermon, *International Conference on Machine Learning*, 2018, pp. 2796–2804.

50 K. Tran, W. Neiswanger, J. Yoon, Q. Zhang, E. Xing and Z. W. Ulissi, *Mach. Learn.: Sci. Technol.*, 2020, **1**, 025006.

51 M.-H. Laves, S. Ihler, J. F., L. A. Kahrs and T. Ortmaier, *arXiv*, 2021, preprint, arXiv:2104.12376, DOI: **10.48550/arXiv.2104.12376**.

52 C.-I. Yang and Y.-P. Li, *J. Cheminf.*, 2023, **15**, 13.

53 Y. Chung, I. Char, H. Guo, J. Schneider and W. Neiswanger, *arXiv*, 2021, preprint, arXiv:2109.10254, DOI: **10.48550/arXiv.2109.10254**.

54 M. Abadi, P. Barham, J. Chen, Z. Chen, A. Davis, J. Dean, M. Devin, S. Ghemawat, G. Irving, M. Isard, *et al.*, *Osdi*, 2016, 265–283.

55 F. Pedregosa, G. Varoquaux, A. Gramfort, V. Michel, B. Thirion, O. Grisel, M. Blondel, P. Prettenhofer, R. Weiss, V. Dubourg, *et al.*, *J. Mach. Learn. Res.*, 2011, **12**, 2825–2830.

56 Y. Zhao, L. Wang, Y. Tian, R. Fonseca and T. Guo, *International Conference on Machine Learning*, 2021, pp. 12707–12718.

57 A. Wojtuch, T. Danel, S. Podlewska and Ł. Maziarka, *J. Cheminf.*, 2023, **15**, 81.

58 P. Pernot, *J. Chem. Phys.*, 2022, **157 14**, 144103.

59 J. Busk, P. B. Jørgensen, A. Bhowmik, M. N. Schmidt, O. Winther and T. Vegge, *Mach. Learn.: Sci. Technol.*, 2021, **3**, 015012.

60 L. van der Maaten and G. E. Hinton, *J. Mach. Learn. Res.*, 2008, **9**, 2579–2605.